%% file: draft_Feb2026.tex
\PassOptionsToPackage{table}{xcolor}
\documentclass{article}
\usepackage{amsmath}
\usepackage{amsthm}
\usepackage{float} 
\usepackage{geometry}

\usepackage{graphicx} 
\usepackage{booktabs}
\usepackage{tikz}
\usepackage[most]{tcolorbox} 
\usepackage{xcolor} 
\usepackage{subcaption}
\usetikzlibrary{positioning, arrows.meta}
\usepackage{amssymb}
\usepackage{natbib}
\usepackage{array}
\usepackage{longtable}
\usepackage{threeparttable}
\usetikzlibrary{arrows.meta, calc, fit, backgrounds}
\usepackage{newtxtext,newtxmath}

\usepackage{hyperref}

\newtcolorbox{instructionbox}{colback=blue!5!white,colframe=blue!75!black,title=Instructions}

\title{Modeling Story Expectations: \\A Generative Framework using LLMs}

\author{Hortense Fong 
    \and 
    George Gui
    \and
    Bo Yang\thanks{Columbia Business School, Hortense Fong (hf2462@gsb.columbia.edu), George Gui (zg2467@gsb.columbia.edu), Bo Yang (BYang30@gsb.columbia.edu). We thank researchers whose comments and suggestions have greatly improved the paper.
    We also thank Yuting Deng, Raelynn Li, Riccardo Risi, Angela Qianya Wang, and Duoji Jiang for exceptional research assistance.}}
\date{May 2026}
\begin{document}

\maketitle

\begin{abstract}
Consumers’ engagement with stories is shaped by their expectations about what will happen next, yet modeling these forward-looking beliefs over unstructured narrative content has remained challenging. We develop a framework that uses large language models to approximate consumers' story expectations. Our method generates multiple imagined story continuations from a pre-trained LLM and extracts interpretable, theory-motivated features from these continuations, such as emotion and narrative path features. We propose two complementary validation procedures suited to different data availability: a survey-based approach that compares LLM-derived expectations to human-reported beliefs, and a rational-expectations approach that compares them to actual story outcomes. Applying the framework to both survey data collected in a controlled lab setting and observational data from an online reading platform, we find that LLM-derived expectations correlate with human-reported beliefs as well as actual story continuations along all features studied. In both settings, forward-looking expectations are associated with reader engagement above and beyond features of the content already consumed. Our framework provides a scalable method for modeling consumer beliefs about narrative content, with implications for content creation, platform strategy, and the study of narrative media.
\end{abstract}

\newpage
\section{Introduction}

Consumers around the world spend a large part of their free time consuming narrative media, whether in the form of TV shows, films, books, podcasts, or narrative-driven social content. For example, U.S. consumers spend around 12 hours a day consuming media, and at least half is on media typically presented in narrative  (i.e., story) form.\footnote{\url{https://www.marketingcharts.com/television-236479}} Despite the significant time allocation and size of the entertainment and media market, it is notoriously difficult for book publishers and television and movie studios to predict what content will be successful \citep{waldfogel2018digital}. We posit that part of the challenge is that story engagement decisions depend in part on consumers' expectations, which have historically been challenging to elicit or model. This research develops a framework that leverages large language models (LLMs) to model consumers' forward-looking expectations for narrative content. 

Modeling consumers' forward-looking expectations allows us to explore audiences' preferences over anticipated content. How do consumers' expectations affect engagement decisions? Why are some stories page turners? Why do some stories generate a lot of buzz? These are questions of great interest to authors and screenwriters, as well as publishers and platforms. For content creators, anticipating consumers' expectations and knowing what affects engagement can help creators design more engaging stories. For publishers and platforms, this information can help improve marketing decisions, such as pricing and advertising.\footnote{For example, emotional state is known to moderate consumer response to ads \citep{goldberg1987happy} and certain emotional states, such as hope, dread, and fear, depend on consumer expectations.}

However, modeling consumers' forward-looking expectations over narratives is challenging because expectations are rarely observed, and approximating them typically requires surveys of the target audience or modeling assumptions. Neither is straightforward in the context of stories. Surveying expectations along one or two dimensions is feasible, but eliciting a full distribution of possibilities over how a story might unfold is cost-prohibitive. Modeling assumptions is also difficult. In structured settings, empirical modelers often assume rational expectations and model beliefs about a future low-dimensional outcome, such as next period's price, as a conditional distribution given past low-dimensional outcomes, such as past prices. For stories, the analogous move would require modeling how the distribution of possible continuations, a high-dimensional object, depends on the story so far, which is also a high-dimensional object. There is no obvious starting point for doing so.

To overcome these challenges, we introduce a novel framework that takes advantage of the vast knowledge base of LLMs to model consumers' expectations in stories. Our framework is made up of four components: story imagination, feature extraction, validation, and analysis. In the story imagination component, we use a pre-trained LLM to generate story continuations. Trained using the text of thousands of books,\footnote{It is speculated that GPT-4 is trained on over 125,000 books. Source:  https://aicopyright.substack.com/p/has-your-book-been-used-to-train} LLMs can predict many probable story continuations from some initial text. For example, providing the LLM with the text from the first chapter, we can ask it to predict the text of the next chapter. Furthermore, we can ask it to generate not just one continuation but a distribution of continuations. The feature extraction component converts moments of this distribution into interpretable features that are motivated by theories from narratology, psychology, and marketing. Our proposed method complements these theories and their associated feature-engineering approaches by examining not only how the original features are associated with engagement, but also how expectations of those features help explain engagement. The validation component assesses whether the previous two components can reasonably model consumers' expectations. We propose two validation procedures. The first is a standard survey-based approach that compares LLM-derived expectations against human-reported beliefs. The second is a rational expectations-based approach that compares LLM-derived expectations against actual story continuations. Finally, the analysis component presents a framework to interpret the association between expectation and engagement, highlighting the challenge of studying it causally and the need for extensive exploration of different story elements before making causal claims. The first three components are about modeling expectations, while the fourth component generates a set of hypotheses attempting to link the narrative features with engagement. 

We apply our method to two complementary empirical settings using survey data and observational data. The survey setting provides a controlled environment with cleaner data and direct measures of participants' beliefs, allowing us to validate whether our expectation-modeling procedure provides a reasonable proxy for consumers' next-chapter expectations. The observational setting captures real stories and organic engagement patterns, allowing us to assess whether the procedure is consistent with rational expectations. We find that current LLMs provide imperfect but informative proxies for consumers' beliefs. LLM-derived expectations correlate with human-reported beliefs as well as actual story continuations across all features studied. Forward-looking expectations are also associated with engagement in both settings. In the survey, expected arousal predicts stated interest in continuing and the decision to read the next chapter, whether expectations are elicited from readers or derived from imagined continuations. In the observational data, LLM-derived expectations of all fifteen features we examine are significant predictors of the realized chapter features, and expected valence, arousal, speed, and circuitousness are associated with voting and commenting rates. Furthermore, we use LLM-derived expectations to assess the sensitivity of these focal relationships to other narrative features, illustrating how the framework can sharpen subsequent causal questions and inform survey design. 

The rest of the paper is organized as follows. Section \ref{section:literature} overviews the relevant literature we build from and contribute to. Section \ref{section:method} introduces the general framework, detailing our approach to modeling expectations in unstructured narrative data. We then demonstrate our method using both survey and observational data in Sections \ref{section:empirical_results_survey} and \ref{section:empirical_results_observational}, respectively. We discuss alternative ways to operationalize our framework in Section \ref{section:alternative_approaches}. Finally, we discuss our main findings in Section \ref{section:discussion} and conclude in Section \ref{section:conclusion}. We hope that by addressing the challenges of modeling expectations in unstructured narrative data, this paper can serve as a starting point for empirical researchers to explore and formulate new hypotheses about how consumer beliefs drive their behavior in contexts involving stories.

\section{Relevant Literature}\label{section:literature}

\input{Sections/relevant_lit}

\section{Method}\label{section:method}

\input{Sections/methods}

\section{Empirical Application with Survey Data}\label{section:empirical_results_survey}
\input{Sections/survey_data}

\subsection{Application: Feature Exploration}\label{section:hypothesis_generation}

\input{Sections/hypothesis_generation}

\section{Empirical Application with Observational Data}\label{section:empirical_results_observational}

\input{Sections/empirical_application_observation}

\section{Alternative Implementations}\label{section:alternative_approaches}

\input{Sections/alternative_approaches}

\section{Discussion}\label{section:discussion}

\input{Sections/discussion}

\section{Conclusion}\label{section:conclusion}

\input{Sections/conclusion}

\bibliography{reference}
\bibliographystyle{apalike}

\newpage
\setcounter{table}{0}
\renewcommand{\thetable}{\Alph{section}\arabic{table}}
\setcounter{figure}{0}
\renewcommand{\thefigure}{\Alph{section}\arabic{figure}}

\appendix 

\section{Appendix: Story Stimuli Generation for Survey}\label{appendix:story_generation}

\input{Sections/appendix_story_generation}

\section{Appendix: Sensitivity of Number of Imagined Continuations} \label{appendix:num_responses}

\input{AppendixSections/sensitivity_N}

\section{Appendix: Survey Details}\label{appendix:qualtrics_survey}

\input{AppendixSections/survey_details}

\section{Appendix: Survey Validation Robustness Checks}\label{appendix:survey_validation_robustness}

\input{AppendixSections/survey_validation_robustness}

\section{Feature Extraction Implementation Details}\label{appendix:feature_extraction}

\input{AppendixSections/feature_extraction}

\newpage
\appendix
\setcounter{section}{0}
\setcounter{figure}{0}
\setcounter{table}{0}
\section{Web Appendix: Extraction of 11 Narrative Dimensions} \label{appendix:extraction_narrative}

\input{AppendixSections/extraction_narrative}

\section{Web Appendix: Data Preprocessing for Observational Data} \label{appendix:wattpad_cleaning}

\input{AppendixSections/data_preprocessing}

\section{Web Appendix: Sensitivity of Semantic Path Features} \label{appendix:semantic_sensitivity}
\input{AppendixSections/semantic_sensitivity}

\section{Web Appendix : Embedding PCA Implementation Details} \label{appendix:embeddings}

\input{AppendixSections/embedding_pca}

\end{document}

%% file: Sections/relevant_lit.tex
This paper relates to three streams of literature: 1) narrative features, 2) consumers' forward-looking expectations, and 3) using large language models (LLMs) for market research. Researchers across marketing, narratology, computer science, and psychology have made progress in extracting features from story text that predict narrative engagement or success \citep{piper2021narrative}. However, these methods primarily describe content already consumed rather than what readers expect will happen next. Much of this work focuses on emotion, building on the premise that entertainment is fundamentally an emotional experience \citep{tan2008entertainment}. \citet{berger2012makes} demonstrate that positive emotions and high-arousal emotions in news stories causally drive sharing behavior. \citet{berger2023holds} extend this to sustained attention, finding that the effects of emotional language on continued reading are driven by the degree to which discrete emotions evoke arousal and uncertainty. Others find that the emotional flow of narratives predicts engagement and/or success \citep{nabi2015role,maharjan2018letting,knight2024narrative,knight2025building}. 
Beyond emotion, researchers have also extracted features related to semantic path \citep{toubia2021quantifying}, story elements \citep{eliashberg2007story,shachar2025sell}, character types \citep{bamman2014bayesian}, character traits \citep{toubia2019extracting}, and antecedents to narrative transport \citep{van2014extended}. 
Our paper extends this literature by adding a forward-looking dimension: modeling what readers expect will happen next, alongside what has already happened in the story. Our proposed method complements existing feature construction methods by applying the feature extraction not only to the actual story text, but also to imagined story continuations generated by a large language model. 

This framework rests on the premise that what readers expect, not just what they have read, shapes engagement. In economics and marketing, models of consumer decision-making recognize that consumers form beliefs about future states from currently available information and act on those expectations. \citet{friedman1957permanent}
assumes that consumers make consumption decisions based not only on current income but also on beliefs about future income. \citet{manski2004measuring} develops a framework for measuring expectations using surveys by eliciting subjective probability distributions, arguing that such data can be used to validate assumptions models make about expectations. Extensive empirical research rests on this assumption. Structural models have estimated dynamic behavior in domains including durable goods purchases and dynamic pricing \citep{rust1987optimal, nair2007intertemporal, gowrisankaran2012dynamics}, consumer stockpiling \citep{hendel2006measuring, ching2020identification}, brand choice and learning under uncertainty \citep{erdem1996decision}, and agents' strategic effort allocation \citep{misra2011structural}.
Together, this literature emphasizes that incorporating expectations of future states is crucial for accurately modeling dynamic behavior.

Regarding forward-looking beliefs in the context of stories, \citet{ely2015suspense} write down a model of suspense and surprise and derive the optimal information revelation strategy over time to maximize expected suspense or surprise. \citet{simonov2023suspense} operationalize these ideas using structured game score data in an online streaming context, demonstrating that suspense, but not surprise, about which team will win predicts viewing behavior. These empirical approaches all require well-defined state spaces, leaving open how to model expectations over unstructured narrative content, where the space of possible continuations is difficult to pre-specify. We extend the expectations framework to unstructured narratives by generating belief distributions directly from narrative text. 

Because LLMs were designed to generate probable text continuations conditional on partial input \citep{radford2019language}, they offer a way to generate distributions of imagined continuations. However, their alignment with human narrative expectations remains untested, and studies exploring whether LLMs can capture human beliefs, judgments, and behaviors have produced mixed results. Even before modern LLMs, neural language models could generate coherent story passages that humans preferred over baselines \citep{fan2018hierarchical}. However, it remains unclear whether LLMs can capture how authors write. \citet{tian2024narratives} finds that narratives generated by GPT-4 lack tension and are more positive than narratives generated by humans. The results are also mixed in other domains. \citet{bybee2023surveying} finds that LLMs can recover belief distributions over economic outcomes that closely match existing survey measures. \citet{halawi2024approaching} find that LLMs are approaching human-level performance on forecasting tasks. \citet{filippas2024large} show that LLMs behave like humans in a number of economics experiments. Several studies find that LLMs are useful for conducting market research \citep{brand2023using,li2024frontiers,lee2024generative,arora2025hybrid,boughanmi2025reviews,timoshenko2026transforming}. At the same time, \citet{goli2024capturing} find that LLMs do not capture human intertemporal preferences. \citet{gao2025take} find that LLMs do not exhibit human-like behaviors in the 11-20 money request game. \citet{peng2025mega} show that digital twins (LLM-based models of real individuals) generate answers that only correlate weakly with human responses and demonstrate systematic bias. Overall, results on LLM-human alignment are mixed, with agreement depending on task specificity and elicitation method. We contribute to this literature by proposing two strategies to validate our framework for modeling story expectations. First, we propose a rational expectations-based strategy,\footnote{Rational expectations is the hypothesis that consumers' subjective beliefs coincide, on average, with the true data-generating process.} examining whether LLM-generated beliefs are, on average, unbiased predictors of actual narrative text. Second, we compare LLM-derived expectations against human-reported beliefs collected through survey. Bringing these three streams together creates a new opportunity for scalable market research on creative content.

%% file: Sections/methods.tex
Our approach leverages the ability of LLMs to approximate consumers' forward-looking beliefs about stories. LLMs are trained on enormous datasets encompassing diverse narratives across various genres, cultures, and time periods \citep{radford2019language}. This expansive training allows them to capture complex patterns in storytelling that would be difficult to model explicitly. In our context, this means the LLM can generate plausible story continuations that reflect the nuanced ways in which narratives typically unfold, mirroring the expectations formed by consumers with broad exposure to stories. We combine LLMs' generative capabilities with theories from psychology, narratology, and economics to extract interpretable features from the LLM-imagined story continuations.

We overview our proposed framework for modeling expectations in stories in Figure~\ref{fig:methodology_flowchart}. The approach consists of four main components: 1) imagined story generation, 2) feature extraction, 3) expectation validation, and 4) learning how the features relate to engagement.

\begin{figure}[h]
    \centering
    \caption{Overview of Proposed Method}
    \label{fig:methodology_flowchart}
    \begin{tikzpicture}[
    node distance=1cm and .74cm,
    box/.style={
        rectangle,
        draw,
        rounded corners,
        minimum width=2.5cm,
        minimum height=2.2cm,
        align=center,
        font=\small,
        fill=blue!10
    },
    smallbox/.style={
        rectangle,
        draw,
        rounded corners,
        minimum width=2.2cm,
        minimum height=1.7cm,
        align=center,
        font=\footnotesize,
        fill=green!10
    },
    validationbox/.style={
        rectangle,
        draw,
        rounded corners,
        minimum width=2.8cm,
        minimum height=2.2cm,
        align=center,
        font=\small,
        fill=yellow!10
    },
    arrow/.style={
        ->,
        >=stealth,
        thick,
        draw=gray
    },
    label/.style={
        font=\footnotesize,
        align=left
    }
]

\node[box] (input) at (-4.5,-2) {
    \textbf{Story Input}
    \\ Story Text
};

\node[smallbox, right=of input] (llm) {
    \textbf{LLM}
};

\node[box, right=of llm] (stories) {
    \textbf{Imagined} \\
    \textbf{Continuations}
    \\ Imagined Continuation 1
    \\ \vdots
    \\ Imagined Continuation N
};

\node[box, right=of stories] (features) {
    \textbf{Feature} \\
    \textbf{Extraction}
};

\node[box, right=of features] (prediction) {
    \textbf{Engagement}
};

\node[validationbox, below=1cm of features] (validation) {
    \textbf{Validation }
    \\ Based on Rational
    \\ Expectations or Survey
};

\draw[arrow] (input.east) -- (llm.west);
\draw[arrow] (llm.east) -- (stories.west);
\draw[arrow] (stories.east) -- (features.west);
\draw[arrow] (features.east) -- (prediction.west);

\draw[arrow] (features.south) -- (validation.north);
\draw[arrow] (validation.west) -| (llm.south);

\end{tikzpicture}
\end{figure}

\subsection{Imagined Story Continuation Generation}

Our starting point is a piece of narrative content, such as the first few chapters of a book. LLMs are trained to predict the next word given a set of preceding words. More specifically, language models estimate the conditional probability of seeing word $w_i$, given all the previous words: $p(w_i|w_1,...,w_{i-1})$. We provide the LLM story text as input (e.g., chapters $1$ through $t$) and ask it to generate the next part of the story (e.g., chapter $t+1$). Given the large amounts of story data used to train LLMs, there is alignment between their training data and the task of generating story continuations. This approach aims to model the process of a representative consumer anticipating potential story developments.

We generate multiple imagined continuations of the story to approximate potential continuations an average consumer might anticipate. We generate multiple continuations for two reasons. First, doing so allows us to obtain a distribution of continuations, from which we can measure not only the mean but also higher-order moments. Second, LLM text generation contains inherent noise, even when temperature is 0,\footnote{\url{https://martynassubonis.substack.com/p/zero-temperature-randomness-in-llms}} so averaging over multiple continuations reduces LLM-generation noise. There are three key design choices in this step: the prompt, the temperature, and the number of continuations. The prompt sets the context for the LLM. Our starting point will be to use a simple prompt:
\begin{quote}
    \textbf{System Prompt:} ``You are an average reader.'' \\
    \textbf{User Prompt:}``Continue this story with the next chapter: [story text]. Next chapter:''
\end{quote}
where [story text] represents the chapters read so far. In this paper, we model beliefs over the next chapter $t+1$. The framework applies to continuations for any number of chapters after $t$. Next, the temperature affects the creativity of the generated text, and we use the default temperature of 1 to balance text diversity and coherence with the input text.\footnote{The temperature parameter scales the probabilities of the next word that the model can select from. A temperature of 0 generates more deterministic output, while a temperature of 2.0 generates more unpredictable output. We found that high temperature values generate incoherent text strings.} Finally, we generate 50 imagined continuations per chapter. A sensitivity analysis in Appendix \ref{appendix:num_responses} shows that feature distributions stabilize around this number.

The choices above define a baseline implementation of the method rather than a fixed recipe. The validation procedures in Section~\ref{section:validation} provide a principled way to refine each. With additional data such as survey responses, richer prompts can also be developed and evaluated against the simple baseline. For example, prompts could condition on reader characteristics to capture heterogeneous beliefs.

\subsection{Feature Extraction}

With multiple imagined story continuations per chapter, the next question is how to quantify and interpret the unstructured story text. We approach this question in two steps: 1) we extract from the text predefined features that have been proposed in the literature to be associated with engagement or narrative success and 2) we calculate consumers' expectations of those features. The core idea is that if consumers care about those features in the text they have already consumed, then they may also care about those same features in what they believe is yet to come.

As discussed in Section \ref{section:literature}, the literature has identified various sets of features based on existing text that are associated with engagement or narrative success, such as emotion and semantic path. A key benefit of these features is that they are relevant to a wide range of stories. However, researchers and practitioners interested in particular genres could use more specific features with our framework. Let $i$ represent the focal book, $t$ the focal chapter, $n$ the imagined continuation number with $N$ capturing the total number of imagined continuations per chapter. Let $\widetilde{story}_{i,t+1}^{n}$ be the imagined continuation for the next chapter and $f$ the representation of a story that reflects existing theory as well as a hypothesis that a researcher is interested in testing. For example, text emotion has been found to be associated with virality \citep{berger2012makes} and a researcher may want to study whether expectations of future emotion in a story help generate word-of-mouth. The researcher would use $f$ to represent the consumed content and imagined continuations as emotion features. We define $\widetilde{z}_{i,t+1}^n$ to be the extracted feature from imagined continuation $n$ (i.e., $\widetilde{z}_{i,t+1}^n=f(\widetilde{story}_{i,t+1}^{n})$) and $z_{i,t}$ to be the extracted feature from chater $t$.

To calculate the expectation of each feature for each chapter, we average the features extracted from the imagined continuations. The expectation of feature $z$ is:
\begin{equation*}
\mathbb{E}^t_n[\widetilde{z}_{i,t+1}^n] = \frac{1}{N} \sum_{n=1}^{N} \widetilde{z}_{i,t+1}^n,
\label{eq:expectation}
\end{equation*}
\noindent which represents the average expected future state of the narrative for feature $z$.

Beyond the first moment, the framework can also be used to calculate other moments of interest, such as variance $\mathrm{Var}_n^t[\widetilde{z}_{i,t+1}^n] = \frac{1}{N-1}\sum_{n=1}^{N}\left(\widetilde{z}_{i,t+1}^n - \mathbb{E}_n^t[\widetilde{z}_{i,t+1}^n]\right)^2$, deviation $\mathrm{Dev}^t_n[z_{i,t},\widetilde{z}_{i,t}^n] = z_{i,t} - \mathbb{E}^{t-1}_n[\widetilde{z}_{i,t}^n]$, interactions $\mathbb{E}^t_n[\widetilde{z}_{i,t+1}^n \widetilde{x}_{i,t+1}^n] = \frac{1}{N} \sum_{n=1}^{N} \widetilde{z}_{i,t+1}^n\widetilde{x}_{i,t+1}^n$, and suspense and surprise \citep{ely2015suspense}. For this paper we primarily focus on empirically evaluating the first moment (i.e., expectation), and we discuss the challenges \citep{bisbee2024synthetic, deng2026examining} and opportunities to studying other moments in Section \ref{section:survey_validation} and Appendix \ref{appendix:survey_validation_robustness}. 

\subsection{Validation of Expectations}\label{section:validation}

Our proposed framework for modeling story expectations is only useful insofar that the extracted features proxy actual expectations. We propose two ways to validate our expectations modeling method: 1) using the actual story continuations, which serve as another approximation of consumer expectations if we assume rational expectations, and 2) a more standard approach using stated beliefs based on surveys with human respondents. These validation approaches enable researchers to continuously validate new LLM models that may be used for story continuation generation and feature extraction.

\subsubsection{Assuming Rational Expectations}

In the context of narratives, assuming rational expectations means assuming that consumers who understand how writers typically craft stories form beliefs that align with the objective distribution of story continuations. Rational expectations is an assumption commonly made in the structural dynamic discrete choice literature \citep{rust1987optimal,erdem1996decision,hitsch2006empirical, dube2010tipping}. Using the actual story continuations to proxy consumer expectations is easy to implement but relies on rational expectations being a reasonable assumption.

We formalize this validation procedure below. Let $P(story)$ be the probability distribution over all possible stories. This distribution represents the space of narratively coherent and realistic stories that could potentially be created by humans. Any actual story can be viewed as a single realization from the joint distribution: $story_{i} \sim P(story)$. From the distribution of stories, we can derive a conditional distribution of stories that represents the probability distribution of possible continuations given a specific narrative context (e.g., chapters $1$ through $t$ in story $i$). The actual next chapter can be viewed as a single realization from the conditional distribution: $story_{i,t+1} \sim P(story_{i,t+1} | story_{i,1:t})$ where $story_{i,1:t}$ represents the content from chapters $1$ through $t$ in story $i$. Let $B(story_{i,t+1} | story_{i,1:t})$ represent the representative consumer's subjective belief distribution about the next chapter. Let $\widetilde{story}_{i,t+1}^{n}$ denote the $n$th imagined continuation (i.e., next chapter) generated by an LLM based on chapters $1$ through $t$ of story $i$. Our method tries to approximate the subjective belief distribution, such that $\{\widetilde{story}_{i,t+1}^{n}\}_{n=1}^N$ forms an empirical distribution $\hat{B}(story_{i,t+1} | story_{i,1:t})$ for the next chapter.

Then, assuming rational expectations, consumers who understand how writers typically craft stories should form beliefs that align with the objective writer-generated distributions (i.e., $B(story_{i,t+1} | story_{i,1:t}) = P(story_{i,t+1} | story_{i,1:t})$). We assess whether the expectations implied by consumers' subjective belief distribution $\hat{B}(story_{i,t+1} | story_{i,1:t})$ align with the expectations implied by the objective distribution $P(story_{i,t+1} | story_{i,1:t})$. We operationalize this by representing the actual chapter text and imagined continuations using $f$ and regressing the actual features $f(story_{i,t+1})$ on the expected features $\mathbb{E}[f(\widetilde{story}_{i,t+1})\mid story_{i,1:t}]$). For example, if we were interested in the expected valence of the next chapter, we would extract the valence of the imagined and actual next chapters, and then regress each chapter's actual valence on its average LLM-based valence. An intercept of 0 and coefficient of 1 would suggest that the expectation modeling method perfectly captures consumers' expectations under the rational expectations assumption. A statistically significant positive coefficient would indicate the method captures meaningful signal about consumers' expectations. The same procedure can be used for the other moments of the feature distribution as well.

\subsubsection{Survey}

While the rational expectations-based validation approach is easy to implement and does not require additional data, it relies on the assumption of rational expectations, which may not hold. The survey-based validation provides a direct benchmark for the subjective belief distribution our method seeks to approximate. Although it requires costly data collection, it avoids imposing the rational expectations assumption. 

Without cost constraints, we would ideally ask survey respondents to write out their beliefs over possible story continuations to obtain the joint distribution over story features that consumers care about. This would allow us to understand consumer preferences over any given feature or combinations of features. However, a full distribution over text is cost prohibitive, so we instead propose a validation procedure that focuses on theoretically motivated dimensions and elicit belief distributions following \citet{manski2004measuring}.

As before, $f$ maps a story continuation to features of interest (e.g., valence). We define an outcome space for each feature and elicit beliefs over that space. For concreteness, suppose feature $f$ induces a finite set of mutually exclusive outcomes $\mathcal{K}_f=\{1,\ldots,K_f\}$. For example, valence can be characterized by five ordered bins, ranging from very negative (1) to very positive (5). For each story context $story_{i,1:t}$ and feature $f$, the survey asks respondents to report a probability vector $\mathbf{s}_{i,t+1,f} \equiv \left(s_{i,t+1,f}(1),\ldots,s_{i,t+1,f}(K_f)\right)$, where $s_{i,t+1,f}(k)$ is a respondent's stated probability that outcome $k$ will occur in the next chapter. We interpret the distribution of $\mathbf{s}_{i,t+1,f}$ across all respondents as an empirical proxy for the feature-level beliefs implied by $B(story_{i,t+1} | story_{i,1:t})$.
As was done for the rational expectations validation procedure, we assess alignment between survey beliefs and LLM-implied beliefs by regressing survey respondents' expectations on our LLM-generated expectations $\mathbb{E}[f(\widetilde{story}_{i,t+1})\mid story_{i,1:t}]$. The same can be done for other distribution moments too. 

The two procedures provide a strategy for validating how well our belief approximation method captures consumers' expectations about how stories continue. They can also help refine the choice over the prompt, temperature, and LLM model.

\subsection{Explaining Engagement}\label{section:explaining_engagement}

\input{Sections/explaining_engagement.tex}

%% file: Sections/explaining_engagement.tex
If the LLM-based framework generates valid expectations that approximate consumers' expectations, we can then use the LLM-derived expectations to better understand narrative engagement. To illustrate the value of modeling a consumer's foward-looking beliefs, consider a consumer reading chapter $t$ of book $i$ who decides whether to continue. Let $U_{i,t}$ denote her expected utility from continuing past chapter $t$ that is determined by the story read so far $story_{i,t}$. Because $story_{i,t}$ is a high-dimensional text object, researchers typically need to choose a $K$-dimensional feature set $\mathbf{f}: \text{Text} \to \mathbb{R}^K$ that serves as a \textit{sufficient representation} \citep{veitch2020adapting, egami2022make,feder2022causal} of the story: the consumer's beliefs about the next chapter and her engagement decision depend on the chapter text only through these features. Let $\mathbf{z}_{i,t} = \mathbf{f}(story_{i,t})$ denote the vector of realized features in the story up through chapter $t$, and let $\boldsymbol{\alpha}, \boldsymbol{\beta} \in \mathbb{R}^K$ denote the corresponding coefficient vectors. Consider a linear expected utility model:\footnote{We adopt the linear form for clarity. It extends to nonlinear or dynamic utility, without changing the distinction between preferences over realized and anticipated content.}
\begin{equation}
U_{i,t} \;=\; \boldsymbol{\alpha}^\top \mathbf{z}_{i,t} \;+\; \boldsymbol{\beta}^\top \mathbb{E}\!\left[\mathbf{f}(\widetilde{story}_{i,t+1}) \,\middle|\, \mathbf{f}({story_{i,t}})\right] \;+\; \varepsilon_{i,t},
\tag{Expected Utility}
\label{eq:utility}
\end{equation}
where $\varepsilon_{i,t}$ represents an idiosyncratic taste shock. The consumer's expectation of next-chapter features depends on the story so far, represented by $\mathbf{f}(story_{i,t})$, as well as 
how her beliefs form, summarized by a $K \times K$ matrix $\Gamma$ whose entries capture how the current story shapes her expectations about the the next chapter:
\begin{equation}
\mathbb{E}\!\left[\mathbf{f}(\widetilde{story}_{i,t+1}) \,\middle|\, \mathbf{f}(story_{i,t})\right] \;=\; \Gamma\, \mathbf{z}_{i,t}.
\tag{Belief Formation}
\label{eq:belief_formation}
\end{equation}

The model implies several parameters of interest. To make them concrete, consider a representation $\mathbf{f}(story_{i,t}) = (V_{i,t}, A_{i,t}, \xi_{i,t})^\top$, where $V$ denotes valence, $A$ arousal, and $\xi$ collects all remaining elements of the story, and $\boldsymbol{\alpha} = (\alpha_V, \alpha_A, \alpha_\xi)^\top$, $\boldsymbol{\beta} = (\beta_V, \beta_A, \beta_\xi)^\top$, and $\Gamma_{j,k}$ denotes the entry of the belief-formation matrix giving the effect of current feature $k$ on the expectation of next-chapter feature $j$. Take emotional arousal, which captures how high or low energy the chapter is, as the running example. The \textit{total effect of current arousal} on engagement bundles every channel through which current arousal can move the consumer's utility from continuing:
\begin{align*}
\delta_A \;&=\; \alpha_A \;+\; (\Gamma^\top \boldsymbol{\beta})_A =\underbrace{\alpha_A}_{\text{direct effect}} \;+\; \underbrace{\Gamma_{V,A}\,\beta_V}_{\text{via expected valence}} \;+\; \underbrace{\Gamma_{A,A}\,\beta_A}_{\text{via expected arousal}} \;+\; \underbrace{\Gamma_{\xi,A}\,\beta_\xi}_{\text{via expected }\xi}.
\end{align*}
The first term is the direct effect: how much a high-arousal chapter raises engagement on its own. The remaining terms run through belief formation: an arousing chapter shifts what the consumer expects to read next (the column $\Gamma_{\cdot,A}$), and she has preferences over each anticipated dimension (the entries of $\boldsymbol{\beta}$). Figure \ref{fig:belief_formation} illustrates the various channels through which current arousal affects the consumer's expected utility. $\delta_A$ collapses these into a single number, mixing structural preferences with belief formation. Thus, the total effect of current arousal is not a pure preference parameter. It combines the consumer’s taste for current arousal with the way current arousal shifts expectations.

\input{Figures/explaining_engagement_tikz}

Modeling expectations to recover $\beta_A$ is valuable because preference over expectations are micro-founded economic primitives that can generalize to cases when the belief formation process associated with a story changes. In many applications, preferences over realized and anticipated content may be more portable than the belief-formation process itself. For example, consumers' beliefs may change as they gain genre familiarity. Without being able to measure or approximate $\Gamma$, researchers cannot understand readers' preferences over expectations. 

Modeling expectations also delivers a prediction benefit by amplifying the value of existing theory. For example, prior work suggests that valence and arousal are important drivers of engagement. If readers care about current valence and arousal, they may also care about expected future valence and arousal. Including expected valence and arousal as features therefore captures, indirectly, how other story dimensions $\xi$ feed into the consumer's beliefs about what comes next in terms of valence and arousal. We illustrate this idea in Figure \ref{fig:belief_formation}. Expected next-chapter valence and arousal depend on current $\xi$ through the off-diagonal entries $\Gamma_{V,\xi}$ and $\Gamma_{A,\xi}$. Although $\xi$ may itself be high-dimensional and difficult to model explicitly, the LLM-generated continuations summarize its predictive content in part through expected valence and arousal. Including expectations as features can therefore improve predictive power without requiring an explicit model of every contributing $\xi$ dimension or sacrificing interpretability. 

The model clarifies why estimating the effect of anticipated arousal is difficult even in a randomized experiment due to the unstructured nature of the data \citep{wei2025unstructured}. Suppose a researcher randomly assigns readers to manipulated stories, elicits their expectations about next-chapter arousal, and measures subsequent engagement. Random assignment makes exposure to the manipulated stories exogenous, but the manipulation may change many story features at once. A regression of engagement on expected arousal alone therefore does not generally recover $\beta_A$. Instead, the coefficient on expected arousal reflects both the structural preference for anticipated arousal and bias from omitted current and expected features that co-vary with expected arousal: $\text{plim }\hat{b}=\beta_A+\sum_{X\in{A,V,\xi}}\alpha_X\pi_X+\sum_{X\in{V,\xi}}\beta_X\pi_{\mathbb{E}X}$, where each $\pi_X$ is the cross-story slope of feature $X$ on expected arousal. 
The estimated slope thus bundles the structural preference $\beta_A$ with the effects of other story features that co-vary with expected arousal. Understanding these channels requires measuring or approximating expected next-chapter content along multiple dimensions, which LLM-generated continuations can provide at scale.

More generally, the coefficient on any feature in any specific representation is best read as a partial association, holding the other included features fixed, which is how we are going to interpret the regression coefficients for the remainder of the paper. While not causal, these coefficients can inform researchers on what additional data to collect or analysis to conduct. When $\xi$ is correlated with the included current or expected features, that coefficient absorbs the corresponding share of the $\xi$ effect as omitted-variable bias. Expanding the set of features included, whether through richer feature extraction or additional survey items, both shrinks $\xi$ and closes one specific channel of bias on the previously included coefficients.

%% file: Figures/explaining_engagement_tikz.tex
\begin{figure}[htbp]
\centering
\caption{Unstructured Content and Expected Utility}
\label{fig:belief_formation}
\resizebox{0.75\textwidth}{!}{%
\begin{tikzpicture}[
    font=\small,
    box/.style={
        draw,
        rounded corners,
        align=center,
        minimum width=1.15cm,
        minimum height=0.75cm,
        fill=white
    },
    outerbox/.style={
        draw,
        rounded corners,
        inner xsep=0.35cm,
        inner ysep=0.35cm,
        fill=gray!3
    },
    llmbox/.style={
        draw,
        rounded corners,
        inner xsep=0.50cm,
        inner ysep=0.65cm,
        fill=gray!1
    },
    arrow/.style={-{Latex[length=2mm]}, thick},
    dashedarrow/.style={-{Latex[length=2mm]}, thick, dashed},
    dottedarrow/.style={-{Latex[length=2mm]}, thick, dotted},
    lab/.style={fill=white, inner sep=1.2pt}
]

\node[align=center] at (0,3.0) {Realized content};
\node[align=center] at (4.7,3.0) {Expected future content};
\node[align=center] at (9.2,3.0) {Expected utility};

\node[box] (A)  at (0,1.6) {$A$};
\node[box] (V)  at (0,0.5) {$V$};
\node[box] (X1) at (0,-0.6) {$\xi_1$};
\node[align=center] (Xdots) at (0,-1.3) {$\vdots$};
\node[box] (Xn) at (0,-2.0) {$\xi_n$};

\node[box] (EA)  at (4.7,1.6) {$\mathbb{E}[A]$};
\node[box] (EV)  at (4.7,0.5) {$\mathbb{E}[V]$};
\node[box] (EX1) at (4.7,-0.6) {$\mathbb{E}[\xi_1]$};
\node[align=center] (EXdots) at (4.7,-1.3) {$\vdots$};
\node[box] (EXn) at (4.7,-2.0) {$\mathbb{E}[\xi_n]$};

\begin{scope}[on background layer]
\node[
    draw,
    rounded corners,
    fill=gray!1,
    minimum width=2.85cm,
    minimum height=4.65cm,
    anchor=center
] (llmgroup) at (2.35,-0.20) {};

\node[outerbox, fit=(A)(V)(X1)(Xdots)(Xn)] (actualgroup) {};
\end{scope}

\node[font=\small] at ($(llmgroup.south)+(0,0.28)$) {Survey/LLM};

\node[box, minimum width=2.0cm, minimum height=0.9cm] (U) at (9.2,0.5) {$U$};

\draw[arrow] (A.east) -- 
    node[lab, pos=0.66, above] {$\Gamma_{A,A}$} 
    (EA.west);

\draw[arrow] (A.east) -- 
    node[lab, pos=0.64, above, sloped] {$\Gamma_{V,A}$} 
    (EV.west);

\draw[arrow] (A.east) -- 
    node[lab, pos=0.62, below, sloped] {$\Gamma_{\xi,A}$} 
    (EX1.west);

\draw[dottedarrow] (X1.east) to[bend left=18]
    node[lab, pos=0.25, above, sloped] {}
    (EA.west);

\draw[dottedarrow] (X1.east) to[bend left=8]
    node[lab, pos=0.30, below, sloped] {$\Gamma_{V,\xi_1}$}
    (EV.west);

\draw[dottedarrow] (Xn.east) to[bend right=30]
    (EA.west);

\draw[dottedarrow] (Xn.east) to[bend right=18]
    (EV.west);

\draw[arrow] (EA.east) -- 
    node[lab, pos=0.48, above, sloped] {$\beta_A$} 
    (U.north west);

\draw[arrow] (EV.east) -- 
    node[lab, pos=0.50, above] {$\beta_V$} 
    (U.west);

\draw[arrow] (EX1.east) -- 
    node[lab, pos=0.50, below, sloped] {$\beta_\xi$} 
    (U.south west);

\coordinate (alphaStart) at (actualgroup.south east);
\coordinate (alphaMidA)  at (4.9,-3.45);
\coordinate (alphaMidB)  at (7.4,-3.35);
\coordinate (alphaEnd)   at ($(U.south west)+(-0.05,-0.25)$);

\draw[dashedarrow]
    (alphaStart) .. controls (alphaMidA) and (alphaMidB) ..
    node[lab, pos=0.72, below, sloped, yshift=-4pt] {$\boldsymbol{\alpha}$}
    (alphaEnd);

\end{tikzpicture}%
}
\end{figure}

%% file: Sections/survey_data.tex
We first apply our proposed expectation-modeling approach in a controlled survey setting. This environment allows us to directly elicit participants' beliefs about upcoming story content, providing a ground-truth benchmark against which to validate our LLM-based expectation measures. We also collect reader engagement to assess whether forward-looking expectations are associated with it, and whether our modeled expectations can detect this relationship. A key benefit of the controlled setting is that consumer selection is absent by design. In Section~\ref{section:empirical_results_observational}, we apply the method to observational data to examine realistic engagement patterns but where such selection is present.

To illustrate our method, we focus on the emotion dimensions of valence and arousal encoded in the text \citep{russell1980circumplex}.\footnote{We distinguish between emotions expressed or encoded in the text and emotions felt by the reader. \citet{mohammad2013crowdsourcing} find that the former is associated with greater agreement between individuals and is therefore what we measure.} Valence captures how positive or negative a passage of text is and arousal captures how high or low energy. These dimensions are well-defined psychological constructs, widely studied in narrative research \citep{berger2021makes,reagan2016emotional,knight2024narrative}, and consistently linked to engagement. Furthermore, forward-looking beliefs are known to shape behavior through the anticipatory emotions they generate \citep{caplin2001psychological}.

\subsection{Survey Data}

We recruit 263 participants via Prolific to read short stories and report their emotional beliefs and engagement. Each participant reads two stories, each consisting of two chapters, yielding 1,052 participant-chapter observations. The stories span five genres (fantasy, literary fiction, romance, science fiction, thriller), with 8 premises per genre for a total of 40 unique story premises. An example of a premise used is: ``Benor Vale, a former pilot, lives in exile on a dusty moon and spends long evenings staring at the wreckage of his ship on the horizon.'' Each premise has a common first chapter and four second-chapter variants that differ in their emotional trajectory (positive/negative valence crossed with high/low arousal), giving 160 unique story variants. This yields a total of 200 unique chapters (40 first chapters and 160 second chapters). We share the details of the procedure used to generate the chapters in Appendix~\ref{appendix:story_generation}. Our goal was to generate story text using a procedure that is easily reproducible and maximizes our ability to measure expectations across the dimensions of valence and arousal. 

After reading each chapter, participants report: (1) the valence and arousal of the text they just read (1--5 scale), (2) their interest in continuing to read (1--5 scale), and (3) a probability distribution over expected valence and arousal for the next chapter (allocating 100 points across five bins). Each unique chapter 1 and chapter 2 combination is read by two to four participants. After the second chapter, participants additionally decide whether to read the third chapter. Participants are told that the third chapter is entirely optional but that if they decide to read it that it will take one to two more minutes. This binary decision serves as a revealed preference measure. We share additional survey details in Appendix \ref{appendix:qualtrics_survey}.  

\subsection{Do Stated Expectations Predict Reader Engagement?}\label{section:survey_expectations_influence_engagement}

We use the survey responses to learn whether forward-looking beliefs are associated with engagement. We test this by regressing reader stated interest to continue reading and their actual decision to continue reading on current valence and arousal alongside the expectation, uncertainty, and/or deviation of valence and arousal.

We estimate the following model:
\begin{equation}
\text{outcome}_{pt} = \mathbf{x}_{pt}'\boldsymbol{\beta} + \alpha_p + \varepsilon_{pt}
\end{equation}

\noindent where $p$ indexes participants and $t$ indexes chapters. The participant fixed effect $\alpha_p$ absorbs unobserved individual heterogeneity in response tendencies. Standard errors are two-way clustered at the participant and chapter level. We measure stated interest in continuing to read after both chapters one and two, and therefore also include chapter number fixed effects when the outcome is interest. We measure the binary decision to continue reading chapter three only after chapter two, and therefore only include chapter two data when the outcome is the continue-to-read decision. We note that deviation can only be measured after chapter two because it requires the expectations generated after reading chapter one. 

The coefficients on the valence and arousal variables capture the association between a story's emotional profile and reader engagement. Because the emotional dimensions may covary with other story attributes, we do not interpret the coefficients as isolated causal effects but instead as a combined association of the measured emotional profile and any correlated story properties. Table~\ref{tab:claim2_expectations} shows the results of regressing interest on the valence and arousal variables collected for both chapters one and two. The stated expectations and uncertainty variables serve as direct measures of consumers' forward-looking beliefs. Adding expectations significantly improves the fit of the model. Stated arousal expectations significantly predict interest in continuing to read. After controlling for the valence and arousal of the chapter that they read (current valence and arousal), readers who anticipate more arousing continuations report greater interest in continuing. Expected valence and uncertainty over valence and arousal, by contrast, are not significant predictors of interest in our survey setting.

\begin{table}[h!]\centering
\small
\begin{threeparttable}
    \caption{Do Stated Expectations Predict Stated Interest?}
    \label{tab:claim2_expectations}
    \input{Tables/survey/e123_survey_uncertainty}
    \begin{tablenotes}[flushleft]
        \footnotesize
        \item \textit{Note:} Two-way clustered standard errors by user and chapter.
        $^{***}p<0.01$; $^{**}p<0.05$; $^{*}p<0.1$.
    \end{tablenotes}
\end{threeparttable}
\end{table}

Table~\ref{tab:claim2c_continue_to_read} reports results for the binary decision of continuing to read chapter three, which we interpret as a revealed-preference measure. Because this was measured after chapter two, only chapter two data is used so the sample size is reduced (N = 526). We estimate the model following equation 1 using a linear probability model. As with the interest outcome, expected arousal predicts the decision to continue reading, while expected valence does not. Absolute deviation and uncertainty also do not predict the decision to continue reading in this setting.

\begin{table}[h!] \centering
\small
\begin{threeparttable}
  \caption{Do Stated Expectations Predict Continue-to-Read Decision?}
  \label{tab:claim2c_continue_to_read}
\input{Tables/survey/e124_ch2_ctr}
  \begin{tablenotes}[flushleft]
  \footnotesize
  \item \textit{Note:} Linear probability model. Two-way clustered standard errors by user and chapter. Ch2 only. $|$Deviation$|$ $=$ $|$actual Ch2 V/A $-$ Ch1 survey expectation$|$. $^{***}p<0.01$; $^{**}p<0.05$; $^{*}p<0.1$.
  \end{tablenotes}
\end{threeparttable}
\end{table}

Altogether, the coefficients in Tables \ref{tab:claim2_expectations} and \ref{tab:claim2c_continue_to_read} suggest that consumers' forward-looking expectations matter for engagement. For both stated interest and decision to continue reading, the models including expectations show a statistically significant improvement in fit. 

\subsection{Survey Validation: Do LLM Expectations Approximate Human Beliefs?}\label{section:survey_validation}

We then validate whether the expectations implied by our LLM-generated continuations correlate with readers' stated expectations. To apply our expectation modeling framework, we generate 50 imagined continuations for each chapter using GPT-4.1-mini\footnote{Model version: \texttt{gpt-4.1-mini-2025-04-14}.} (temperature = 1.0) and extract valence and arousal from each continuation as well as the original story text (temperature = 0.0). Implementation details are in Appendix~\ref{appendix:story_generation}. We compute LLM-based expectations as the average valence and arousal across continuations.

\begin{figure}[!htbp]
    \centering
    \caption{Survey Emotion vs. LLM-derived Emotion}
    \label{fig:claim1_scatter}
    \includegraphics[width=\textwidth]{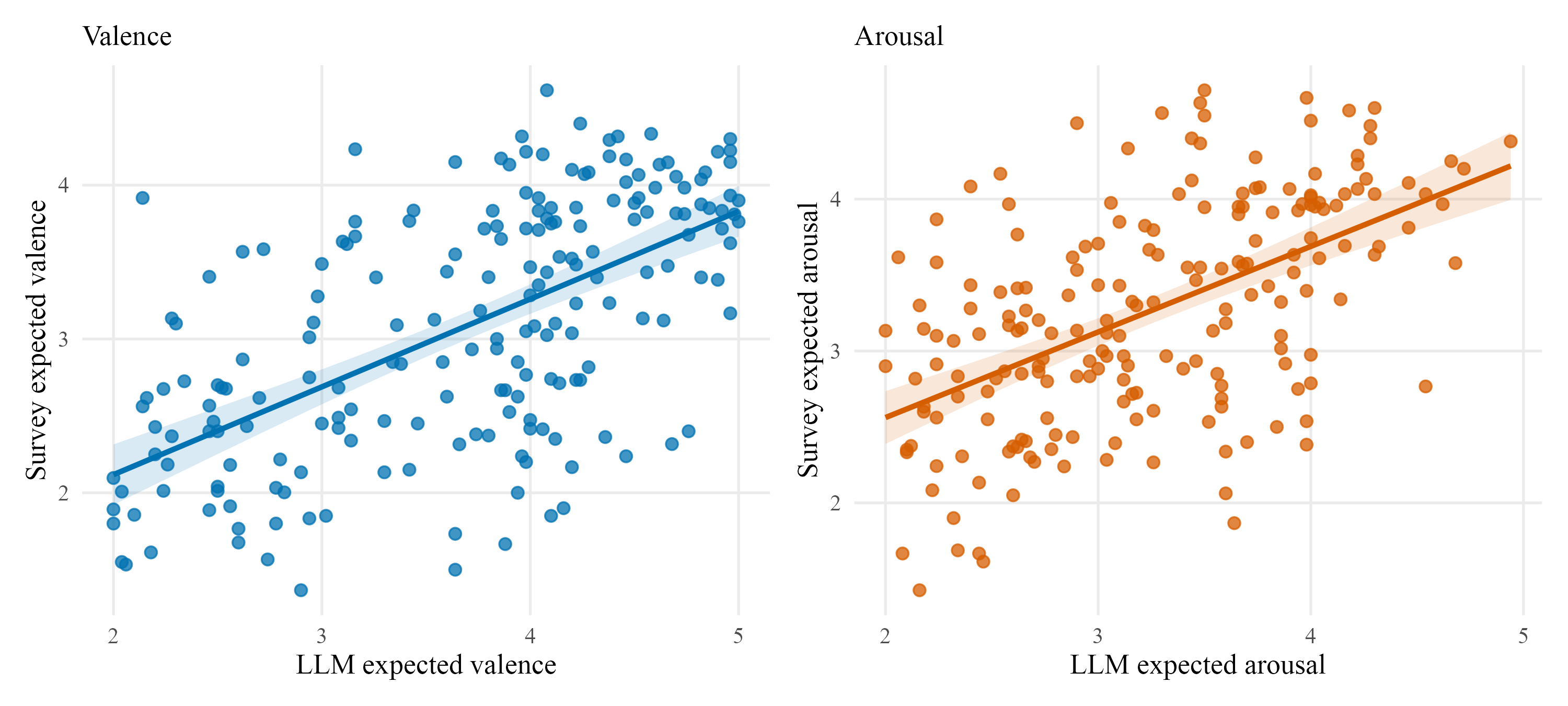}
    \begin{tablenotes}
            \footnotesize
    \item \textit{Note}: Each point is a unique chapter ($N=200$).
    \end{tablenotes}
\end{figure}

Figure~\ref{fig:claim1_scatter} shows the scatter plots of the survey measures against the LLM-based measures. For each chapter, we regress survey-reported expected valence and arousal on LLM-derived expected valence and arousal. Table~\ref{tab:claim1_validation} reports the regression results at the chapter level.

\begin{table}[!htbp] \centering
\small
\begin{threeparttable}
  \caption{LLM Expectations Predict Survey Expectations}
  \label{tab:claim1_validation}
\input{Tables/survey/claim1_validation}
  \begin{tablenotes}[flushleft]
  \footnotesize
  \item \textit{Note:} $^{***}p<0.01$; $^{**}p<0.05$; $^{*}p<0.1$. 
  \end{tablenotes}
\end{threeparttable}
\end{table} 

The slopes in Table~\ref{tab:claim1_validation} are 0.57 for valence and 0.56 for arousal and both significantly positive, indicating that LLM-derived expectations capture meaningful variation in human beliefs. The slopes are below 1, indicating that LLM-based measures are imperfect but informative. The intercepts are positive (0.98 for valence, 1.44 for arousal), indicating a level difference between the two measurement scales. Despite this approximation error, which can lead to attenuation bias, the LLM expectations retain sufficient signal to predict reader interest, as we show below.
This validation holds across both chapters and all five genres, as shown in Figure~\ref{fig:claim1_scatter_genre} and Table \ref{tab:claim1_combined} in Appendix \ref{appendix:survey_validation_robustness}. The relationship also survives controlling for current-chapter valence and arousal (Table~\ref{tab:claim1_validation_controls} in Appendix~\ref{appendix:survey_validation_robustness}).

A well-known limitation of current off-the-shelf LLMs is that they tend to produce overly concentrated distributions relative to humans \citep{bisbee2024synthetic, deng2026examining}. Our validation yields similar findings: when we regress survey uncertainty on LLM-derived uncertainty for valence and arousal, the coefficients are small, statistically insignificant, and explain little of the variation in survey responses (Table \ref{tab:uncertainty_validation} in Appendix \ref{appendix:survey_validation_robustness}). Because of this limitation, this paper focuses on expectations and leaves the modeling of uncertainty to future work.

\subsection{Using LLM Expectations to Understand Reader Engagement}\label{section:llm_expectations_influence_engagement}

Having established that forward-looking expectations are associated with engagement, we ask whether LLM-derived expectations can serve as a scalable proxy. We estimate parallel models replacing survey-elicited expectations with LLM-based measures. Table~\ref{tab:llm_expectations_interest} columns (3) and (4) show the results. We find that the modeled results parallel the survey results. In both the survey and modeled data, current and expected valence do not significantly predict interest, but current and expected arousal do. Furthermore, the coefficient magnitudes are comparable. 

\begin{table}[htbp] \centering
\small
\begin{threeparttable}
  \caption{LLM Expectations and Stated Interest}
  \label{tab:llm_expectations_interest}
\input{Tables/survey/claim2_expectations}
  \begin{tablenotes}[flushleft]
  \footnotesize
  \item \textit{Note:} Two-way clustered standard errors by user and chapter. $^{***}p<0.01$; $^{**}p<0.05$; $^{*}p<0.1$.
  \end{tablenotes}
\end{threeparttable}
\end{table}
Table~\ref{tab:llm_expectations_ctr} reports the results of the binary decision to continue reading chapter three. As with the interest outcome, expected arousal predicts the decision to continue reading in the survey ($p<0.05$) and modeled ($p<0.10$) specifications (columns 2 and 4), while expected valence does not. The coefficient magnitudes are comparable across the two measurement approaches, reinforcing the conclusion that LLM-derived expectations can serve as a scalable proxy for survey-elicited beliefs.

\begin{table}[h!] \centering
\small
\begin{threeparttable}
  \caption{LLM Expectations and Continue-to-Read Decision}
  \label{tab:llm_expectations_ctr}
\input{Tables/survey/claim2c_continue_to_read}
  \begin{tablenotes}[flushleft]
  \footnotesize
  \item \textit{Note:} Linear probability model. Two-way clustered standard errors by user and chapter. Ch 2 only. $^{***}p<0.01$; $^{**}p<0.05$; $^{*}p<0.1$.
  \end{tablenotes}
\end{threeparttable}
\end{table}

%% file: Tables/survey/e123_survey_uncertainty.tex
\begin{tabular}{@{\extracolsep{5pt}}lccc} 
\hline 
\hline  
 & \multicolumn{3}{c}{DV: Stated Interest} \\ 
 \cline{2-4} 
 & Current V/A & + Expectations & + Exp + Uncertainty \\ 
 & (1) & (2) & (3)\\ 
\hline  
 Valence & 0.020 & $-$0.005 & $-$0.005 \\ 
  & (0.021) & (0.026) & (0.026) \\ 
  Arousal & 0.218$^{***}$ & 0.120$^{***}$ & 0.120$^{***}$ \\ 
  & (0.026) & (0.031) & (0.032) \\ 
  Expected valence &  & 0.036 & 0.035 \\ 
  &  & (0.050) & (0.050) \\ 
  Expected arousal &  & 0.204$^{***}$ & 0.206$^{***}$ \\ 
  &  & (0.048) & (0.049) \\ 
  Uncertainty valence &  &  & $-$0.008 \\ 
  &  &  & (0.058) \\ 
  Uncertainty arousal &  &  & 0.036 \\ 
  &  &  & (0.076) \\ 
 \hline  
Chapter number FE & Yes & Yes & Yes \\ 
User FE & Yes & Yes & Yes \\ 
$F$ (vs prev.) &  & 3.32*** & 0.04 \\ 
Observations & 1,052 & 1,052 & 1,052 \\ 
R$^{2}$ & 0.731 & 0.739 & 0.739 \\ 
Adjusted R$^{2}$ & 0.640 & 0.651 & 0.650 \\ 
\hline 
\hline  
\end{tabular} 

%% file: Tables/survey/e124_ch2_ctr.tex
\begin{tabular}{@{\extracolsep{5pt}}lcccc} 
\hline 
\hline  
 & \multicolumn{4}{c}{DV: Continue-to-Read} \\ 
 \cline{2-5} 
 & Current V/A & + Expectations & + Exp + Deviation & + Exp + Dev + Unc \\ 
 & (1) & (2) & (3) & (4)\\ 
\hline  
 Valence & $-$0.0005 & $-$0.008 & $-$0.008 & $-$0.009 \\ 
  & (0.015) & (0.021) & (0.021) & (0.021) \\ 
  Arousal & 0.039$^{**}$ & 0.006 & 0.005 & 0.0004 \\ 
  & (0.015) & (0.020) & (0.022) & (0.022) \\ 
  Expected valence &  & 0.015 & 0.014 & 0.013 \\ 
  &  & (0.027) & (0.027) & (0.028) \\ 
  Expected arousal &  & 0.075$^{**}$ & 0.072$^{**}$ & 0.079$^{**}$ \\ 
  &  & (0.030) & (0.032) & (0.032) \\ 
  $|$Deviation$|$ valence &  &  & 0.020 & 0.011 \\ 
  &  &  & (0.026) & (0.028) \\ 
  $|$Deviation$|$ arousal &  &  & $-$0.002 & $-$0.002 \\ 
  &  &  & (0.026) & (0.026) \\ 
  Uncertainty valence &  &  &  & 0.045 \\ 
  &  &  &  & (0.048) \\ 
  Uncertainty arousal &  &  &  & 0.072 \\ 
  &  &  &  & (0.059) \\ 
 \hline  
User FE & Yes & Yes & Yes & Yes \\ 
$F$ (vs prev.) &  & 3.10** & 0.28 & 2.00 \\ 
Observations & 526 & 526 & 526 & 526 \\ 
R$^{2}$ & 0.721 & 0.727 & 0.728 & 0.732 \\ 
Adjusted R$^{2}$ & 0.439 & 0.448 & 0.445 & 0.449 \\ 
\hline 
\hline  
\end{tabular} 

%% file: Tables/survey/claim1_validation.tex
  \begin{tabular}{@{\extracolsep{5pt}}lcc}
  \hline
  \hline
   & \multicolumn{2}{c}{\textit{Dependent variable:}} \\
  \cline{2-3}
  & Survey expected valence & Survey expected arousal \\
  & (1) & (2)\\
  \hline
   LLM expected valence & 0.571$^{***}$ &  \\
    & (0.053) &  \\
    LLM expected arousal &  & 0.563$^{***}$ \\
    &  & (0.062) \\
    Constant & 0.977$^{***}$ & 1.435$^{***}$ \\
    & (0.200) & (0.205) \\
   \hline
  Observations & 200 & 200 \\
  R$^{2}$ & 0.369 & 0.297 \\
  Adjusted R$^{2}$ & 0.366 & 0.293 \\
  \hline
  \hline
  \end{tabular}

%% file: Tables/survey/claim2_expectations.tex
\begin{tabular}{lcccc}
\hline \hline
 & \multicolumn{2}{c}{Survey} & \multicolumn{2}{c}{Modeled} \\
\cmidrule(lr){2-3} \cmidrule(lr){4-5}
 & Current V/A & + Expectations & Current V/A & + Expectations \\
 & (1) & (2) & (3) & (4) \\
\hline
Valence & 0.020 & -0.005 & 0.012 & 0.022 \\
 & (0.021) & (0.027) & (0.022) & (0.026) \\
Arousal & 0.218*** & 0.120*** & 0.160*** & 0.118*** \\
 & (0.026) & (0.031) & (0.024) & (0.026) \\
Expected valence &  & 0.036 &  & -0.000 \\
 &  & (0.050) &  & (0.058) \\
Expected arousal &  & 0.204*** &  & 0.146** \\
 &  & (0.048) &  & (0.071) \\
\hline
Chapter number FE & Yes & Yes & Yes & Yes \\
User FE & Yes & Yes & Yes & Yes \\
Observations & 1,052 & 1,052 & 1,052 & 1,052 \\
R$^2$ & 0.731 & 0.739 & 0.713 & 0.715 \\
Adjusted R$^2$ & 0.640 & 0.651 & 0.616 & 0.618 \\
\hline \hline
\end{tabular}

%% file: Tables/survey/claim2c_continue_to_read.tex
\begin{tabular}{lcccc}
\hline \hline
 & \multicolumn{2}{c}{Survey} & \multicolumn{2}{c}{Modeled} \\
\cmidrule(lr){2-3} \cmidrule(lr){4-5}
 & Current V/A & + Expectations & Current V/A & + Expectations \\
 & (1) & (2) & (3) & (4) \\
\hline
Valence & -0.000 & -0.008 & -0.005 & -0.011 \\
 & (0.015) & (0.021) & (0.014) & (0.017) \\
Arousal & 0.039** & 0.006 & 0.025 & 0.013 \\
 & (0.015) & (0.020) & (0.018) & (0.020) \\
Expected valence &  & 0.015 &  & 0.058 \\
 &  & (0.027) &  & (0.040) \\
Expected arousal &  & 0.075** &  & 0.077* \\
 &  & (0.030) &  & (0.045) \\
\hline
User FE & Yes & Yes & Yes & Yes \\
Observations & 526 & 526 & 526 & 526 \\
R$^2$ & 0.721 & 0.727 & 0.717 & 0.721 \\
Adjusted R$^2$ & 0.439 & 0.448 & 0.430 & 0.434 \\
\hline \hline
\end{tabular}

%% file: Sections/hypothesis_generation.tex
Having validated the expectation framework on a predefined set of emotional dimensions, we next extend it to other narrative dimensions. As discussed in Section~\ref{section:explaining_engagement}, these approximations may be imperfect, but can nevertheless help researchers refine their causal questions and interpret the empirical relationships they identify.

Causal questions about stories are difficult to study because narratives vary along many dimensions that often move together. For example, changing a story to increase arousal rarely changes arousal alone, but typically changes pacing and tension along with it. By approximating expectations along many dimensions at low cost, our framework helps researchers assess whether these distinctions are empirically consequential and clarify which counterfactual their estimates correspond to. Once the causal question is refined, researchers can more efficiently decide what additional data to collect, based on how each approximated dimension correlates with the focal feature and outcome.

To illustrate, we use the imagined continuations to learn additional narrative dimensions that may be relevant for reader engagement and extract them from both the actual text and imagined continuations. The features are narrative tension, curiosity, emotional complexity, character agency, social conflict, resolution versus complication, cliffhanger intensity, vividness, pacing, character relatability, and cognitive complexity (definitions in Appendix~\ref{appendix:hyp_gen_extraction}). For each dimension, we compute its expected value as the average rating across continuations, following the same procedure used for valence and arousal. Although we do not have human survey responses against which to validate these dimensions directly, we can validate them using the rational-expectations approach discussed in Section~\ref{section:rational_exp_validation}.

\begin{figure}[htbp!]
\centering
\caption{Sensitivity Analysis of V/A Coefficients}
\includegraphics[width=\textwidth]{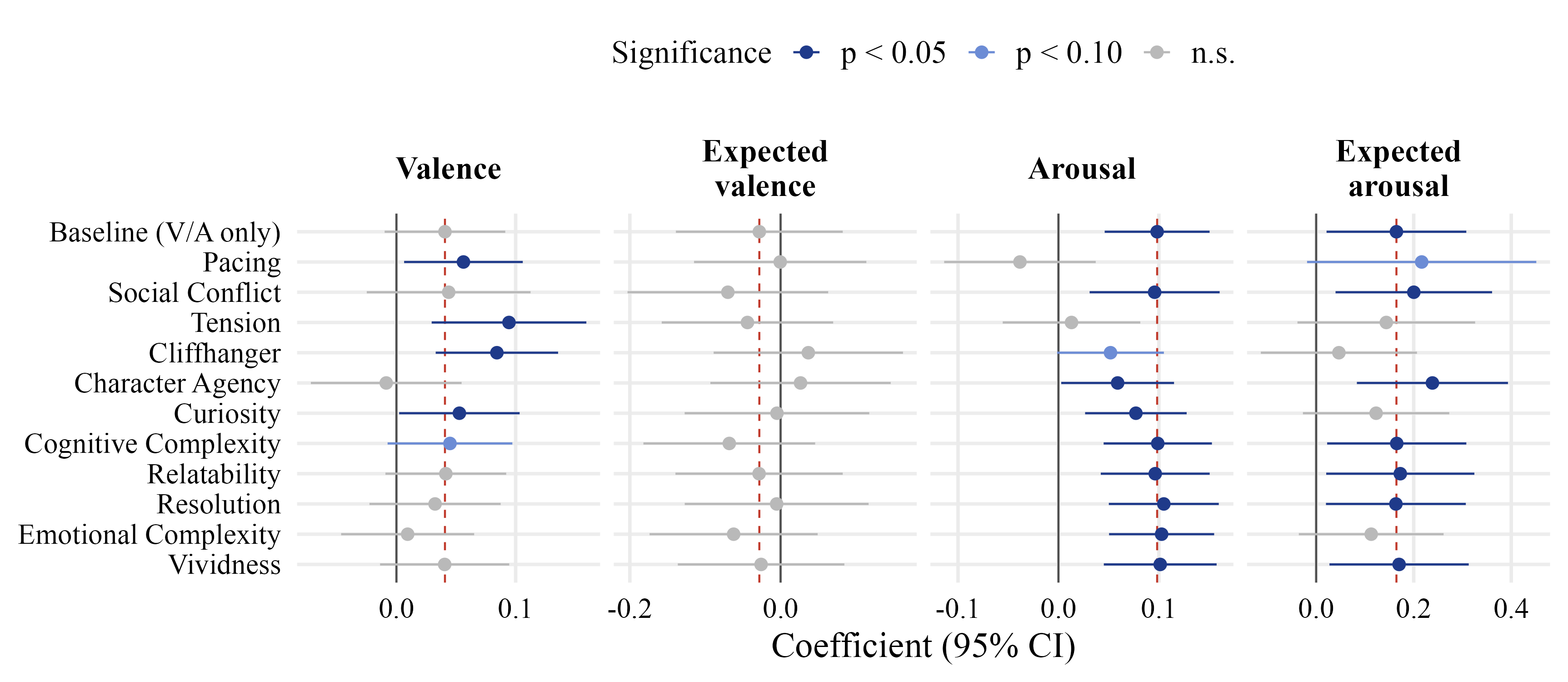}
\vspace{0.5em}
\begin{minipage}{\textwidth}
\footnotesize
\textit{Note}: Sensitivity of the validated coefficients to adding one narrative
dimension at a time. The baseline specification regresses interest on current
and expected valence and arousal, with user and chapter-number fixed effects.
Each row adds the realized and expected ratings of a single dimension to that
baseline. 
\end{minipage}
\label{fig:va_sensitivity}
\end{figure}

We then assess whether adding each dimension, one at a time, changes the valence and arousal coefficients reported in Table~\ref{tab:llm_expectations_interest}. Figure~\ref{fig:va_sensitivity} shows that the four coefficients respond differently to these additions. The current-chapter arousal coefficient is stable and remains significant in nearly every specification, with the exceptions being pacing and tension. By contrast, the expected-arousal coefficient is sensitive to the addition of tension, cliffhanger intensity, curiosity, and emotional complexity. Current-chapter valence becomes significant after controlling for pacing, tension, cliffhanger, and curiosity.

For researchers studying valence and arousal, whether these additional dimensions should be measured depends on the causal question. In particular, the researcher must decide whether a given dimension is a confounder to be held fixed or a mediator that should be allowed to vary. Approximating and exploring these dimensions at low cost helps identify which story elements to prioritize, because both the estimate and its interpretation can be sensitive to their inclusion. For example, a researcher interested in the effect of expected arousal on engagement, regardless of what drives that arousal, need not control for whether the story is at a cliffhanger. A researcher interested instead in the component of expected arousal that arises from semantic content alone learns from this analysis that dimensions such as cliffhanger intensity matter and should consider measuring them in a survey.

%% file: Sections/empirical_application_observation.tex
Next, we apply our proposed framework to observational data from an online reading platform. This data set is noisier and subject to reader self-selection of book choice and continued consumption, and so the empirical patterns may not match those of the survey. However, the observational setting also offers several advantages over the survey setting. First, the stories reflect realistic distributions of narrative content, making it possible to test the rational expectations assumption described in Section~\ref{section:validation}. Second, the engagement measures capture revealed preferences through actual reading, voting, and commenting behavior rather than stated intentions.

\subsection{Online Media Platform Data}

We collect book text with chapter-by-chapter engagement from an online media platform where users read and write stories. The content is posted chapter by chapter, and we observe the read count, vote count, and comment count for each chapter, allowing us to calculate continue-to-read, vote-to-read, and comment-to-read rates. We focus on the vote-to-read and comment-to-read rates as our primary engagement outcomes, and exclude the continue-to-read rate because its data quality is compromised by chapter re-uploads, which reset the read count. 
Vote-to-read and comment-to-read rates do not depend on chapter sequencing and are therefore less affected by chapter re-uploads.

For each book, we collect its title and description, creation date, language, whether the content is for mature audiences, and writer-provided book tags (e.g., ``school", ``drama"). For each chapter, we collect the text, title, date written, comment count, vote count, and read count. We select books that have at least nine chapters and focus on the first nine chapters so that we have a balanced dataset across books, which span a wide range of fiction genres.\footnote{We exclude nonfiction because rational expectations validation may be more likely to succeed in this setting and there is risk of data leakage.}

We use \texttt{gpt-4o-mini-2024-07-18} to generate the imagined story continuations. To alleviate the concern about data leakage (i.e., the engagement metrics and actual story continuations being part of the training data), we only include books published after the training data cutoff for the model.\footnote{\texttt{gpt-4o-mini-2024-07-18}'s training data cutoff is October 2023, and all book chapters in our sample are published December 2023 or later.} This ensures that the content was not used to train the LLM we use to generate the imagined stories. Since our proposed story imagination method relies critically on the quality of the input text, we take several steps to clean the stories, as detailed in Web Appendix \ref{appendix:wattpad_cleaning}. After cleaning, we are left with 8,399 chapters across 937 books.

\subsubsection{Summary Statistics}

Table \ref{table:data_summary_stats} provides summary statistics of our dataset. Compared to more traditional books, which have on average 3,000 to 4,000 words per chapter, the chapters in our data are shorter with an average of 1,968 words. Readers can comment throughout the text, and comments are consolidated at the end of the chapter. Readers can also vote to show their support for a writer and can only vote once per chapter. Our outcome measures of interest are the comment-to-read rate (comment count / read count of chapter) and vote-to-read rate (vote count / read count of chapter). Since the engagement distributions are right-skewed, we winsorize the outcome measures at the 2.5th and 97.5th percentiles and z-score them for the regression analysis.

\begin{table}[htbp]
\centering
\caption{Summary Statistics}
\label{table:data_summary_stats}
\input{Tables/empirical/data_summary_stats}
\end{table}

\subsection{Rational-Expectations Validation: Do LLM Expectations Approximate Actual Story Features?}\label{section:rational_exp_validation}

In addition to the emotion features used in the survey validation, this section includes semantic path features (speed, volume, and circuitousness) and the narrative dimensions learned from the imagined continuations in Section~\ref{section:hypothesis_generation}. The semantic path features can be applied to a wide range of stories and have been shown to predict narrative success \citep{toubia2021quantifying}. Speed captures how quickly the text moves from one idea to another, volume captures how much conceptual ground it covers, and circuitousness captures how roundabout its trajectory is. Following \citet{toubia2021quantifying}, we measure speed, volume, and circuitousness for each chapter and imagined continuation using their word embeddings. Because speed and volume are highly correlated in our data ($r = 0.997$), we retain only speed in the subsequent analysis.\footnote{\citet{toubia2021quantifying} also report a high correlation between speed and volume, $r = 0.88$, in their analysis of TV dialogue.} Additional implementation details are provided in Appendix~\ref{appendix:feature_extraction}.

Following the rational-expectations validation procedure described in Section~\ref{section:validation}, we regress each chapter's realized feature value on the corresponding LLM-based expected value. Table~\ref{tab:claim2_expectations_indirect} presents the results. All features have positive, highly significant coefficients, indicating that LLM-generated expectations are informative predictors of realized chapter content. However, as in the survey validation, the LLM-based expectation measures exhibit bias. The emotion coefficients are similar across the survey and rational-expectations validation procedures: 0.69 versus 0.57 for valence and 0.62 versus 0.56 for arousal. This similarity suggests that rational-expectations validation performs similarly to survey-based validation without requiring costly survey data.

\begin{table}[h!] \centering
\small
\begin{threeparttable}
\caption{Validation based on Rational Expectations}
\label{tab:claim2_expectations_indirect}
\input{Tables/empirical/re_15features_fragment}
\begin{tablenotes}[flushleft]
\footnotesize
\item \textit{Note:} $^{***}p<0.01$; $^{**}p<0.05$; $^{*}p<0.1$. N=8,399.
\end{tablenotes}
\end{threeparttable}
\end{table}

\begin{figure}[htbp!]
\centering
\caption{Interaction Coefficients}
\label{fig:beta-dist-cov-indirect}
\includegraphics[width=0.7\textwidth]{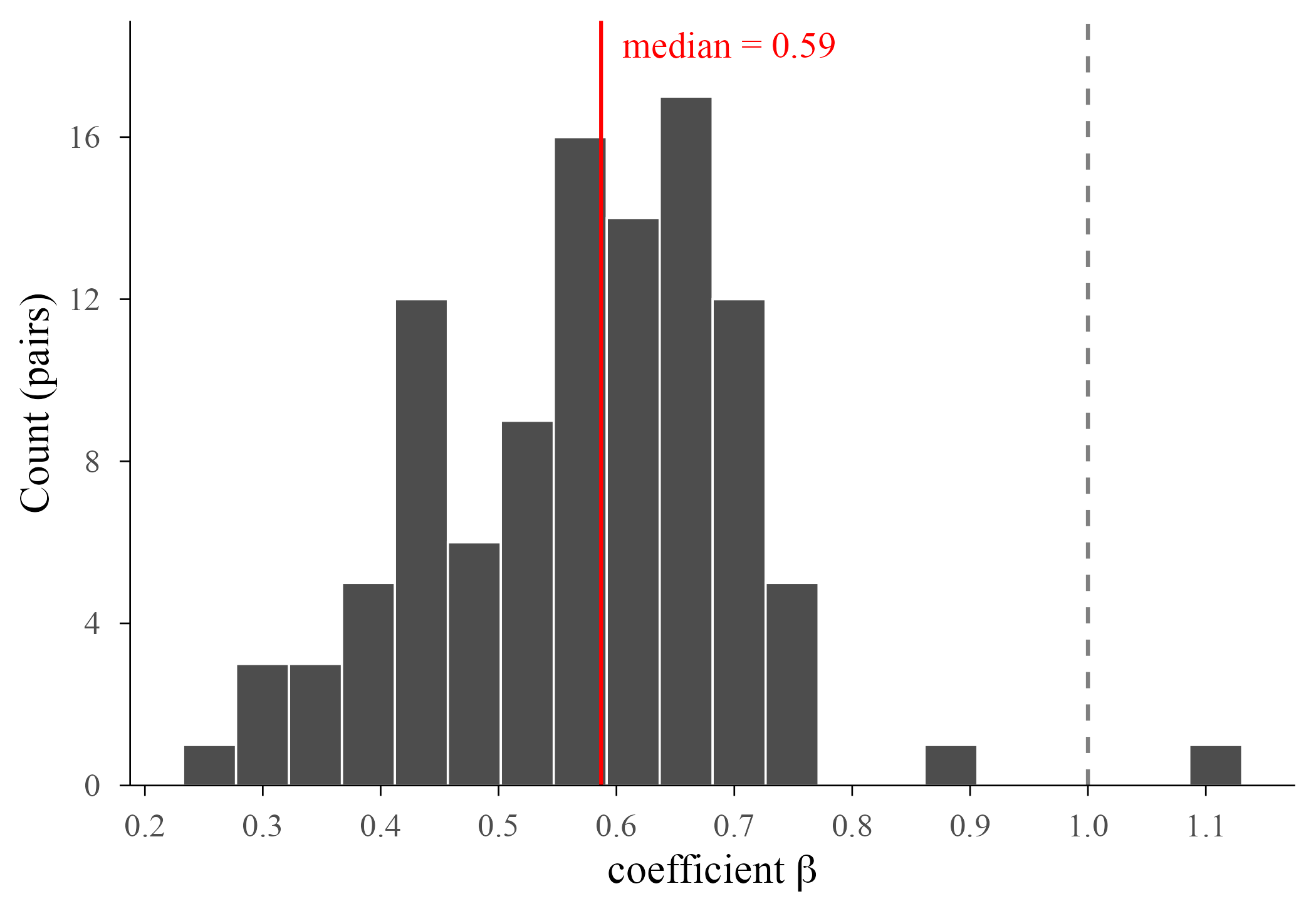}
\par\medskip
\begin{minipage}{0.85\textwidth}
\footnotesize \textit{Note:} Coefficient $\beta$ from $X_{t+1} \cdot Y_{t+1} \sim \mathbb{E}_{\text{indirect}}[X \cdot Y]_t$  across all 105 pairs of the 15 features. $\beta = 1$ is the rational-expectations target (dashed line). Median in red. All 105 coefficients are significant at $p < 0.001$.
\end{minipage}
\end{figure}

In addition, rational-expectations validation allows us to validate features that would be difficult to validate using the survey-based strategy. In particular, it would be challenging to validate measures like interactions and covariances using the survey because they require beliefs about the joint probability. In the survey, we do not measure consumers' beliefs about the joint probability between valence and arousal because of the complexity of the task, which would require assigning probabilities across 25 buckets. By assuming rational expectations, we can compare the pairwise interactions of the features extracted from the imagined continuations with the pairwise interactions of the realized chapters. Figure \ref{fig:beta-dist-cov-indirect} plots the distribution of coefficients from regressing realized next-chapter pairwise interactions on LLM-derived interactions. All of the $\beta$ coefficients are significantly positive and the median coefficient is $0.59$, suggesting our expectations modeling framework captures meaningful variation in human beliefs under the assumption of rational expectations.

\subsection{Which Features Are Associated with Engagement?}

Besides capturing realistic stories, the observational data capture real engagement patterns. Greater engagement in the form of commenting and voting behavior is of interest to platforms, advertisers, and writers as these measures relate to customer retention, ad effectiveness, and product feedback, respectively.

We regress each engagement measure on features of the actual text and expectations from the imagined continuations. We also control for the log of first-chapter read count, which proxies for baseline book popularity, including factors such as author fame and marketing effort. We control for word count because chapter length affects whether consumers finish a given chapter, and include chapter-number fixed effects to absorb systematic differences in engagement across chapter positions, such as reader attrition over the course of a book. Because these observational relationships may reflect consumer heterogeneity and dynamic selection, we interpret the coefficients as platform-specific associations rather than causal effects. Table~\ref{table:engagement_actual_vs_expected} presents the regression results for each set of features separately. Expected valence, arousal, speed, and circuitousness are all significantly associated with voting and commenting. More negative and lower energy next-chapter expectations are associated with higher vote-to-read and comment-to-read rates. Higher speed and lower circuitousness expectations are also associated with greater engagement. As with the survey data, we observe that forward-looking expectations help predict engagement. 

\begin{table}[htbp!]
\centering
\caption{Which Features are Associated with Voting and Commenting?}
\label{table:engagement_actual_vs_expected}
\footnotesize
\input{Tables/empirical/S4_unified_table}
\par\medskip
\begin{minipage}{0.95\linewidth}
\footnotesize \textit{Note:} $^{***}p{<}0.01$; $^{**}p{<}0.05$; $^{*}p{<}0.1$. N = 8,399. Log first-chapter read count and word count included as controls. Chapter number FE included.
\end{minipage}
\end{table}

\begin{figure}[h!]
\centering
\caption{Sensitivity Analysis of V/A Coefficients for Observational Data}
\includegraphics[width=0.95\textwidth]{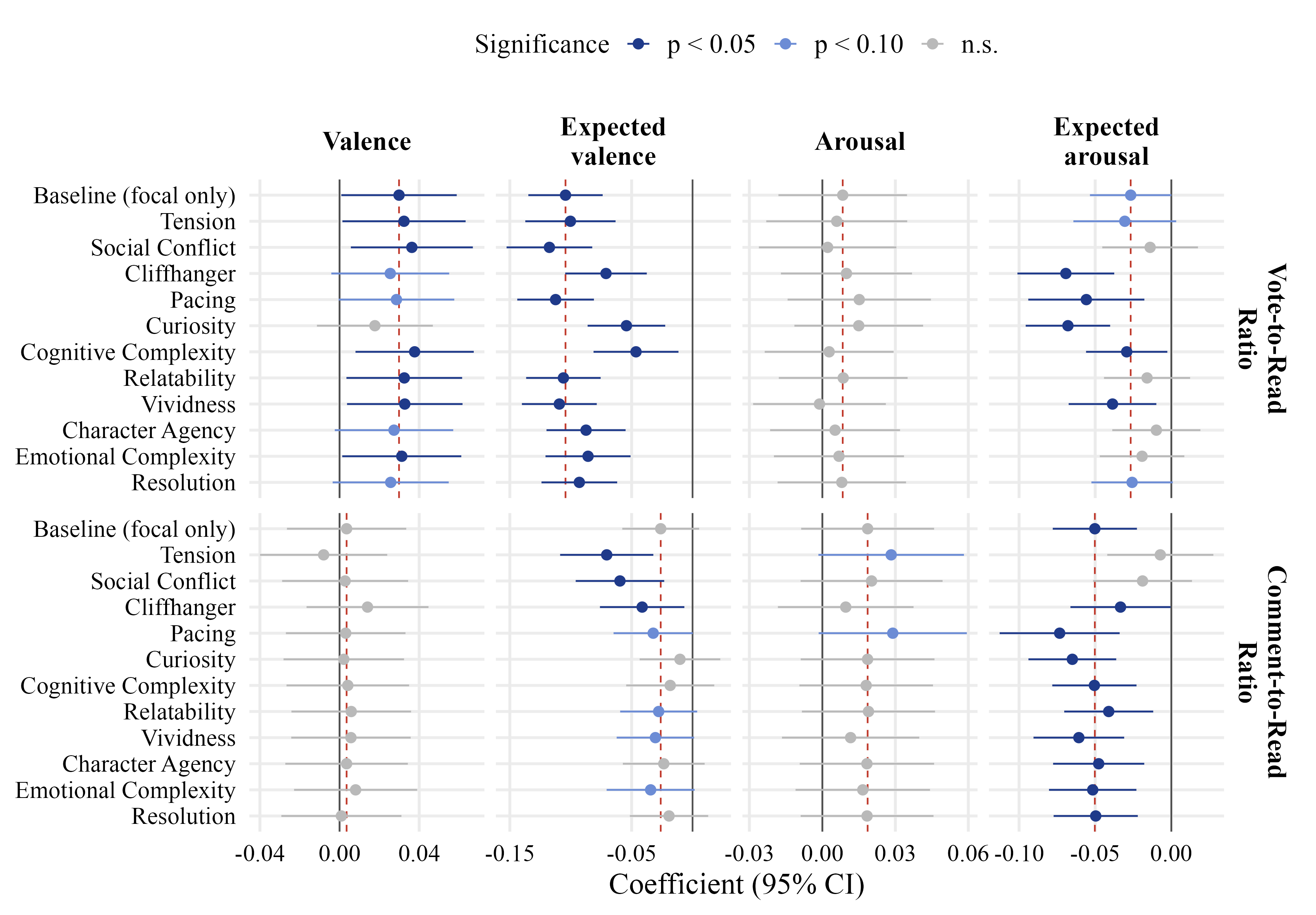}
\vspace{0.5em}
\begin{minipage}{\textwidth}
\footnotesize
\textit{Note}: Sensitivity of the emotion coefficients to adding one narrative dimension at a time. The baseline specification regresses each engagement ratio on current
and expected valence and arousal, with word count and log readership controls and chapter fixed effects.
Each row adds the realized and expected ratings of a single dimension to that
baseline. 
\end{minipage}
\label{fig:obs_va_sensitivity}
\end{figure}

To illustrate how the method can help researchers better understand the relationship between the focal features and engagement, Figure \ref{fig:obs_va_sensitivity} shows how the current and expected valence and arousal coefficients change after including each narrative dimension in the regression, one at a time. The robustness of expected arousal for comment-to-read suggests that readers' anticipation of emotional energy carries signal beyond what dimensions like tension and pacing capture. The greater sensitivity of expected valence for comment-to-read, by contrast, indicates that this association may partly operate through other narrative dimensions.\footnote{Web Appendix \ref{appendix:semantic_sensitivity} provides the results for the semantic path features.} As in Section~\ref{section:hypothesis_generation}, whether to treat these dimensions as confounders to control for or as mediators to allow to vary depends on the researcher's causal question. In this section, we analyze the simplest set of current and forward-looking features, but there are many additional features, such as average past emotion (e.g., average valence and arousal for chapters 1 to $t-1$) and number of narrative reversals \citep{knight2024narrative}, that future work could explore to further enrich our understanding of engagement.

%% file: Tables/empirical/data_summary_stats.tex
\begin{tabular}{lcccccc}
\hline \hline
\textbf{Measure} & \textbf{Mean} & \textbf{Std Dev} & \textbf{Min} & \textbf{P2.5} & \textbf{P97.5} & \textbf{Max} \\
\hline
Number of words           & 1,968 & 1,453 & 200 & 393 & 6,139 & 13,882 \\
Comment count             & 13 & 49 & 0 & 0 & 108 & 1,111 \\
Vote count                & 31 & 157 & 0 & 0 & 171 & 5,338 \\
Read count                & 900 & 4,559 & 1 & 12 & 5,338 & 245,187 \\
Comment-to-read rate      & 0.06 & 0.27 & 0 & 0 & 0.46 & 6.50 \\
Vote-to-read rate         & 0.06 & 0.08 & 0 & 0 & 0.28 & 1.00 \\
\hline \hline
\end{tabular}

%% file: Tables/empirical/re_15features_fragment.tex
\begin{tabular}{@{\extracolsep{5pt}}lccc}
\hline
\hline
 & Intercept & $\beta$ & R$^{2}$ \\
\hline
\multicolumn{4}{l}{\textit{Emotion}} \\
\quad Valence & 0.475$^{***}$ (0.045) & 0.693$^{***}$ (0.012) & 0.298 \\
\quad Arousal & 1.424$^{***}$ (0.045) & 0.615$^{***}$ (0.012) & 0.232 \\
\multicolumn{4}{l}{\textit{Semantic Path}} \\
\quad Semantic Speed & -0.294$^{***}$ (0.019) & 1.285$^{***}$ (0.021) & 0.318 \\
\quad Semantic Circuitousness & 0.015$^{***}$ (0.001) & 0.435$^{***}$ (0.025) & 0.035 \\
\multicolumn{4}{l}{\textit{Learned Features}} \\
\quad Tension & 1.346$^{***}$ (0.045) & 0.670$^{***}$ (0.013) & 0.253 \\
\quad Curiosity & 1.606$^{***}$ (0.045) & 0.607$^{***}$ (0.013) & 0.210 \\
\quad Emotional Complexity & 1.660$^{***}$ (0.051) & 0.559$^{***}$ (0.014) & 0.153 \\
\quad Character Agency & 0.849$^{***}$ (0.079) & 0.721$^{***}$ (0.021) & 0.128 \\
\quad Social Conflict & 1.321$^{***}$ (0.037) & 0.669$^{***}$ (0.012) & 0.260 \\
\quad Resolution & 1.348$^{***}$ (0.044) & 0.350$^{***}$ (0.018) & 0.043 \\
\quad Cliffhanger & 1.327$^{***}$ (0.054) & 0.587$^{***}$ (0.017) & 0.128 \\
\quad Vividness & 1.940$^{***}$ (0.060) & 0.456$^{***}$ (0.018) & 0.073 \\
\quad Pacing & 2.473$^{***}$ (0.066) & 0.382$^{***}$ (0.018) & 0.053 \\
\quad Relatability & 0.218$^{***}$ (0.070) & 0.925$^{***}$ (0.018) & 0.240 \\
\quad Cognitive Complexity & 1.465$^{***}$ (0.053) & 0.623$^{***}$ (0.017) & 0.135 \\
\hline \hline
\end{tabular}

%% file: Tables/empirical/S4_unified_table.tex
\begin{tabular}{@{\extracolsep{4pt}}lcccc}
\hline\hline
 & \multicolumn{2}{c}{Vote-to-Read} & \multicolumn{2}{c}{Comment-to-Read} \\
\cline{2-3} \cline{4-5}
 & Current & + Expectation & Current & + Expectation \\
\hline
\multicolumn{5}{l}{\textit{Panel A. Emotion --- Valence + Arousal, joint multivariate model}} \\
\quad Valence, current & -0.036$^{***}$ (0.012) & 0.032$^{**}$ (0.015) & -0.027$^{**}$ (0.012) & 0.020 (0.015) \\
\quad Valence, expectation & --- & -0.123$^{***}$ (0.015) & --- & -0.089$^{***}$ (0.016) \\
\quad Arousal, current & 0.013 (0.012) & 0.017 (0.013) & -0.020 (0.012) & 0.009 (0.014) \\
\quad Arousal, expectation & --- & -0.045$^{***}$ (0.013) & --- & -0.085$^{***}$ (0.014) \\
\quad $R^2$ & 0.108 & 0.115 & 0.040 & 0.045 \\
\quad Adjusted $R^2$ & 0.106 & 0.113 & 0.038 & 0.044 \\
\hline
\multicolumn{5}{l}{\textit{Panel B. Semantic Path --- Speed + Circuitousness, joint multivariate model}} \\
\quad Semantic Speed, current & 0.023 (0.016) & -0.026 (0.019) & 0.084$^{***}$ (0.016) & 0.010 (0.020) \\
\quad Semantic Speed, expectation & --- & 0.064$^{***}$ (0.014) & --- & 0.095$^{***}$ (0.015) \\
\quad Semantic Circuitousness, current & -0.044$^{***}$ (0.011) & -0.028$^{**}$ (0.011) & -0.055$^{***}$ (0.011) & -0.032$^{***}$ (0.012) \\
\quad Semantic Circuitousness, expectation & --- & -0.047$^{***}$ (0.011) & --- & -0.063$^{***}$ (0.011) \\
\quad $R^2$ & 0.108 & 0.111 & 0.044 & 0.051 \\
\quad Adjusted $R^2$ & 0.106 & 0.110 & 0.043 & 0.050 \\
\hline\hline
\end{tabular}

%% file: Sections/alternative_approaches.tex
To model consumers' forward-looking expectations, we have so far introduced and evaluated a \textit{generative} implementation, where we generate multiple imagined story continuations. We now discuss two alternative implementations that can also operationalize our framework, and discuss the tradeoffs.

\subsection{Direct Elicitation of Features from LLM}

One possible implementation, which we call \textit{direct elicitation}, is to ask the LLM to assign probabilities to possible outcomes for predefined features, mirroring how we elicit beliefs from survey respondents. Using this implementation, we extract emotion features and the 11 narrative dimensions from Section~\ref{section:hypothesis_generation} for both the survey and observational data. The prompt for the narrative dimensions is provided in Web Appendix~\ref{appendix:extraction_narrative}.

Direct elicitation and the generative implementation make different tradeoffs. Direct elicitation is less costly because it does not require generating multiple continuations or extracting features from them. Because its output structurally resembles survey responses, it may better match human-reported beliefs for well-studied features: in the survey-validation comparison (Table~\ref{tab:claim1_validation_direct}), $\beta$ coefficients are closer to one and $R^2$ values are higher than for the generative method. Direct elicitation can therefore be well-suited when the research question targets a predefined set of features and the goal is to approximate stated beliefs without exploring additional dimensions or checking sensitivity to correlated ones.

\begin{figure}[htbp!]
\centering
\caption{Direct Elicitation Interaction Coefficients}
\label{fig:covariance_direct}
\includegraphics[width=0.5\textwidth]{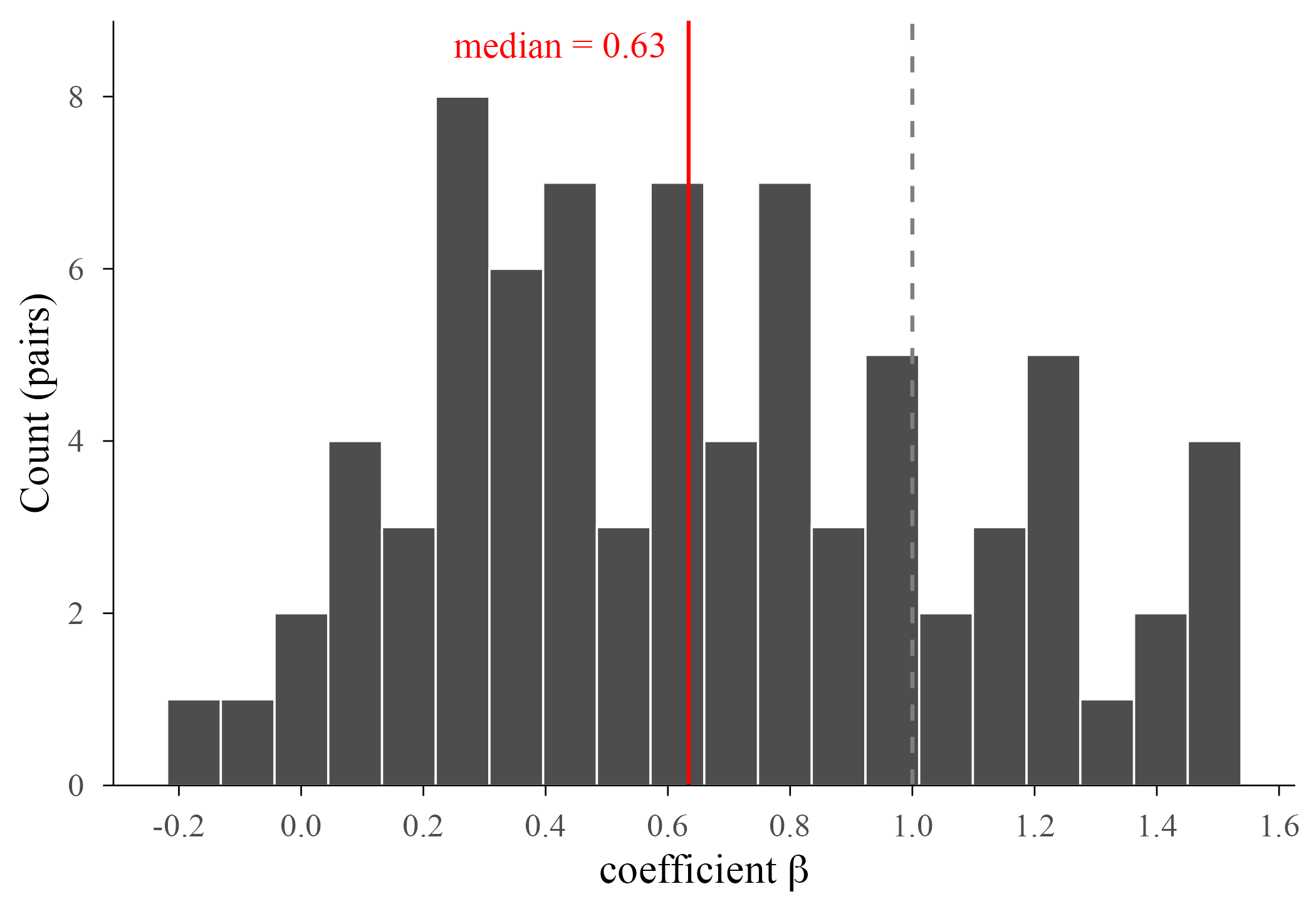}
\par\medskip
\footnotesize \textit{Notes:} All coefficients are significant at $p < 0.001$.
\end{figure}

\begin{figure}[!htbp]
    \centering
    \caption{Comparison of $R^2$ between Implementations (Generative $R^2$ - Direct-elicitation $R^2$)}
    \label{fig:covariances_side_by_side}
    \begin{subfigure}{0.48\textwidth}
        \centering
        \includegraphics[width=\textwidth]{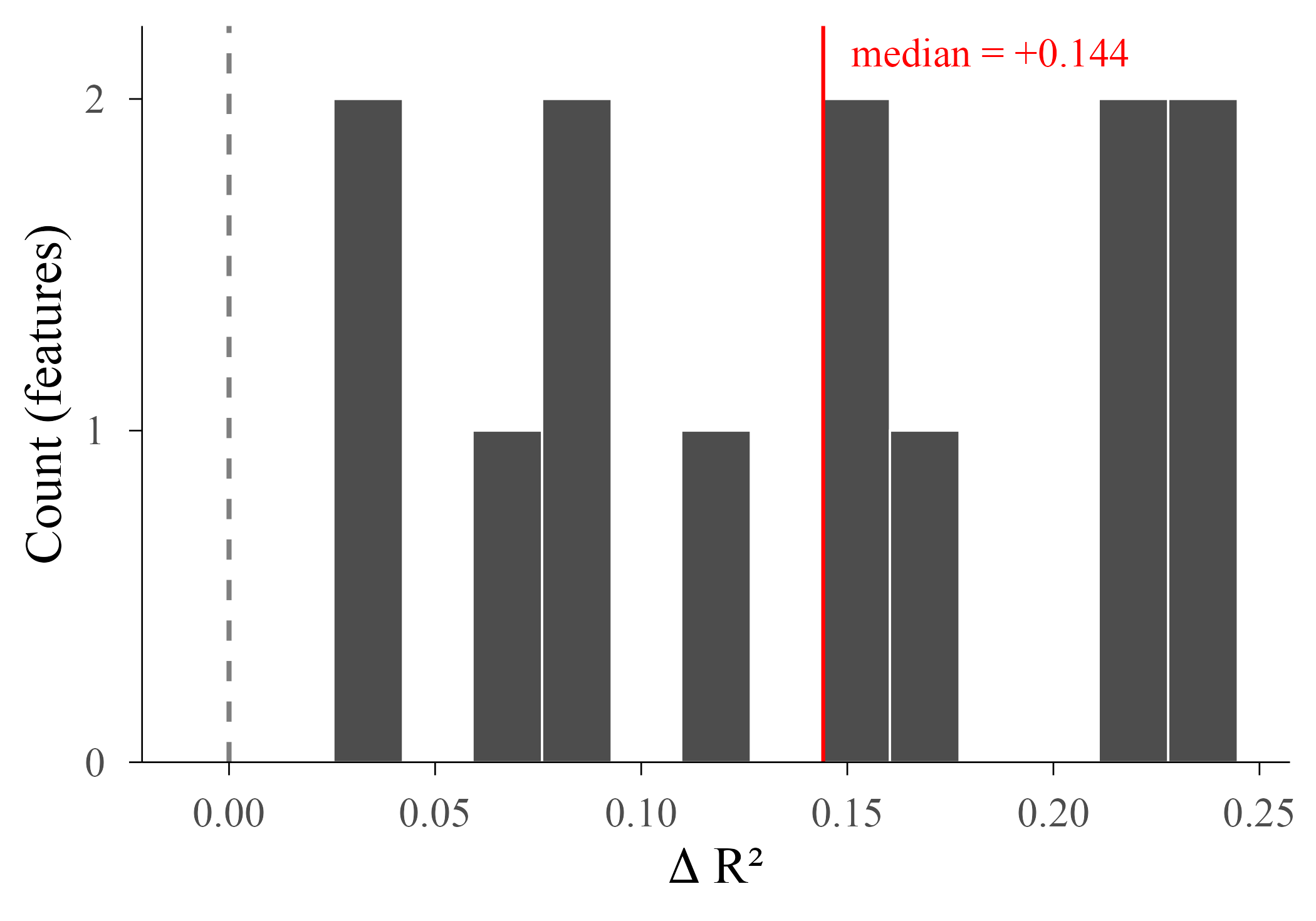}
        \caption{$\Delta R^2$ for Expectations}
        \label{fig:r2_expectations}
    \end{subfigure}
    \hfill
    \begin{subfigure}{0.48\textwidth}
        \centering
        \includegraphics[width=\textwidth]{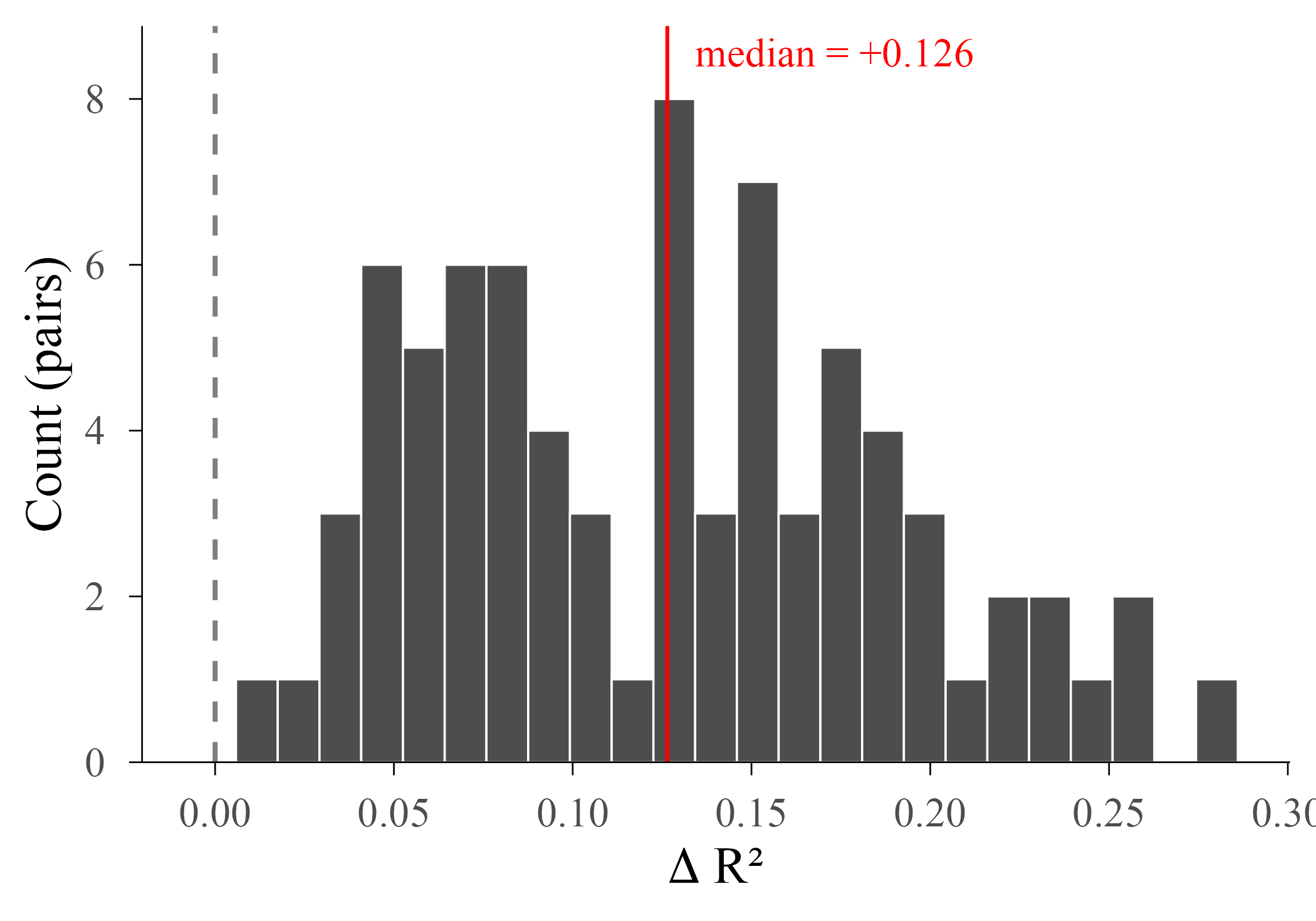}
        \caption{$\Delta R^2$ for Interactions}
        \label{fig:r2_covariances}
    \end{subfigure}
\end{figure}

In contrast, the generative implementation produces continuations as a byproduct of the procedure, which carry several advantages. The continuations give writers a concrete view of what readers may anticipate, can be used to identify additional features beyond those specified a priori (Section~\ref{section:hypothesis_generation}), and can serve as input to feature-extraction methods that require text, such as \citet{toubia2021quantifying}. Because each continuation carries all features simultaneously, the implementation also recovers the joint distribution across feature dimensions, something direct elicitation captures only by separately eliciting probabilities over feature combinations, an exercise that becomes infeasible as the number of bins grows. We find that the generative implementation more closely predicts realized story content for the observational data. While direct elicitation provides a reasonable approximation of realized content, with positive and statistically significant coefficients for nearly all of the expectations features (Table~\ref{tab:claim2_expectations_direct}) and positive interaction coefficients with a median of 0.63 (Figure~\ref{fig:covariance_direct}), the LLM expectations derived from the generative implementation better explain the variation of actual story continuations: Figure~\ref{fig:covariances_side_by_side} shows that the generative $R^2$ exceeds direct-elicitation $R^2$ across all features and feature pairs. Which implementation is preferable therefore depends on the research question.

\begin{table}[!h] \centering
\small
\begin{threeparttable}
  \caption{Direct Elicitation Expectations Predict Survey Expectations}
  \label{tab:claim1_validation_direct}
\input{Tables/survey/claim1_validation_direct}
  \begin{tablenotes}[flushleft]
  \footnotesize
  \item \textit{Note:} $^{***}p<0.01$; $^{**}p<0.05$; $^{*}p<0.1$. 
  \end{tablenotes}
\end{threeparttable}
\end{table}

\begin{table}[htbp!] \centering
  \small
  \begin{threeparttable}
    \caption{Do Direct Elicitation Expectations Predict Actual Story Features?}
    \label{tab:claim2_expectations_direct}
  \input{Tables/empirical/re_13features_fragment}
    \begin{tablenotes}[flushleft]
    \footnotesize
    \item \textit{Note:} $^{***}p<0.01$; $^{**}p<0.05$; $^{*}p<0.1$. N = 8,399.
    \end{tablenotes}
  \end{threeparttable}
  \end{table}

\subsection{Embeddings as Features}

We have so far focused on interpretable features because they are useful to content creators, platforms, and researchers seeking to understand the drivers of engagement. The same framework can in principle be applied when stories are represented using text embeddings rather than predefined interpretable features. The downside is that this approach sacrifices interpretability, and its benefit depends on whether expectations over the embeddings capture meaningful information. Web Appendix~\ref{appendix:embeddings} implements this approach. In the survey data, the embedding-based expectation components do not predict stated interest (Table~\ref{tab:embedding_interest_obs_pca}), possibly due to the limited sample size relative to the number of additional variables. In the observational data, by contrast, the embedding-based expectation components complement the current-chapter components and improve prediction (Table~\ref{tab:09-engagement}).

%% file: Tables/survey/claim1_validation_direct.tex
\begin{tabular}{@{\extracolsep{5pt}}lcc} 
\\[-1.8ex]\hline 
\hline
 & \multicolumn{2}{c}{\textit{Dependent variable:}} \\ 
\cline{2-3} 
& Survey expected valence & Survey expected arousal \\ 
& (1) & (2)\\ 
\hline
 Direct-elicitation expected valence & 0.787$^{***}$ &  \\ 
  & (0.032) &  \\ 
  Direct-elicitation expected arousal &  & 0.773$^{***}$ \\ 
  &  & (0.047) \\ 
  Constant & 0.908$^{***}$ & 0.917$^{***}$ \\ 
  & (0.092) & (0.145) \\ 
 \hline
Observations & 200 & 200 \\ 
R$^{2}$ & 0.755 & 0.582 \\ 
Adjusted R$^{2}$ & 0.754 & 0.580 \\ 
\hline 
\hline
\end{tabular} 

%% file: Tables/empirical/re_13features_fragment.tex
\begin{tabular}{@{\extracolsep{5pt}}lccc}
\hline
\hline
 & Intercept & $\beta$ & R$^{2}$ \\
\hline
\multicolumn{4}{l}{\textit{Emotion}} \\
\quad Valence & 0.915$^{***}$ (0.085) & 0.808$^{***}$ (0.031) & 0.075 \\
\quad Arousal & 3.210$^{***}$ (0.094) & 0.153$^{***}$ (0.030) & 0.003 \\
\multicolumn{4}{l}{\textit{Learned Features}} \\
\quad Tension & 1.991$^{***}$ (0.145) & 0.532$^{***}$ (0.044) & 0.017 \\
\quad Curiosity & 0.213 (0.161) & 1.086$^{***}$ (0.050) & 0.053 \\
\quad Emotional Complexity & 2.385$^{***}$ (0.148) & 0.397$^{***}$ (0.047) & 0.009 \\
\quad Character Agency & 1.281$^{***}$ (0.193) & 0.732$^{***}$ (0.061) & 0.017 \\
\quad Social Conflict & -1.534$^{***}$ (0.163) & 1.564$^{***}$ (0.053) & 0.093 \\
\quad Resolution & 2.743$^{***}$ (0.077) & -0.183$^{***}$ (0.026) & 0.006 \\
\quad Cliffhanger & 0.448$^{***}$ (0.130) & 0.884$^{***}$ (0.042) & 0.051 \\
\quad Vividness & 2.705$^{***}$ (0.193) & 0.247$^{***}$ (0.063) & 0.002 \\
\quad Pacing & 2.180$^{***}$ (0.117) & 0.550$^{***}$ (0.037) & 0.025 \\
\quad Relatability & 2.123$^{***}$ (0.161) & 0.544$^{***}$ (0.051) & 0.013 \\
\quad Cognitive Complexity & 0.034 (0.153) & 1.066$^{***}$ (0.049) & 0.054 \\
\hline
\hline
\end{tabular}

%% file: Sections/discussion.tex
In both the survey and observational data, we find that LLM-generated story continuations correlate with consumers' forward-looking expectations over interpretable dimensions and that these LLM-derived expectations are associated with reader engagement. We discuss the key findings, implications, and limitations below.

The survey validation demonstrates that LLM-derived expectations correlate with human-reported beliefs over story continuations along the dimensions of valence and arousal (Table~\ref{tab:claim1_validation}). The slopes are attenuated relative to unity, indicating that LLM expectations capture the direction and relative ordering of human beliefs but compress the range. In the observational setting, the rational expectations validation confirms that LLM-generated expectations are informative predictors of realized chapter content across all of the features we examine, which includes expectations as well as interactions between pairs of features (Table~\ref{tab:claim2_expectations_indirect}). Together, these results suggest that LLMs trained on large corpora of narrative text can serve as a reasonable starting point for modeling what readers anticipate.

We also find that forward-looking expectations are associated with engagement above and beyond the content readers have already consumed. In the survey setting, expected arousal significantly predicts reader interest after controlling for current valence and arousal (Table~\ref{tab:claim2_expectations}). In the observational setting, the expected emotion and semantic path features predict voting and/or commenting behavior after controlling for their current measures (Table~\ref{table:engagement_actual_vs_expected}). We show the sensitivity of the coefficients to controlling for the narrative dimensions identified using the generated continuations.

Our framework has several implications for research and practice. First, it provides a scalable method for empirically modeling consumer beliefs over narrative content. Because the method can be applied to any text for which story continuations can be generated, it extends the expectations literature beyond structured state spaces to the domain of unstructured narratives. Researchers studying narrative engagement can use our approach to complement existing feature extraction methods by adding a forward-looking dimension.

Second, the finding that expected features are associated with engagement has practical implications for content creators and platforms. For content creators, our results suggest that what readers anticipate matters for engagement, not just what they experience. Authors and screenwriters may benefit from considering how their narrative choices shape reader expectations, such as how chapter endings signal the emotional trajectory or pacing of what is to come. For platforms, the ability to extract expectation-based features at scale could inform recommendation systems, content curation, and ad placement strategies. 

Third, the sensitivity analysis of the coefficients to including additional content-based controls has important implications for researchers seeking to study the causal effect of story features on outcomes of interest. Because story features often move together, an estimate for any given feature depends on which other dimensions are held fixed. Our framework helps researchers diagnose whether omitted dimensions are likely to change the interpretation of a coefficient. LLM-based feature extraction can help refine the causal question before researchers commit to a particular experimental or survey design.

Our approach has several limitations that present opportunities for future research. First, while our validation results are encouraging, LLM-derived expectations are approximations. The attenuated survey-validation slopes suggest that they compress the range of human beliefs, introducing measurement error that can attenuate downstream regression coefficients. Future work could explore fine-tuning LLMs or calibrating the generated distributions to better match human belief distributions. Second, our empirical results are associational rather than causal. The regression coefficients capture the relationship between narrative features and engagement conditional on controls, but they may reflect correlated story attributes rather than the isolated effect of any single dimension. Experimental designs that manipulate reader expectations while holding actual content constant could help establish causal relationships. Third, our current implementation focuses on textual narratives. Many forms of narrative media involve visual and auditory elements that our text-based approach does not capture. For example, \citet{manzoor2023designing} use a music generation model to quantify the unpredictability of music. As multimodal large language models continue to develop, extending our framework to generate and analyze multimodal story continuations is a natural next step. Finally, we model a representative consumer by treating the LLM as an average reader, but in practice, consumers are heterogeneous in their narrative preferences and expectations. Readers with different genre expertise, cultural backgrounds, or reading histories may form systematically different beliefs about how stories will unfold. Future work could explore conditioning the LLM on reader-specific information or modeling heterogeneous expectations across consumer segments.

%% file: Sections/conclusion.tex
This paper introduces a framework for modeling consumers' forward-looking expectations in stories by using large language models. By extracting features from these continuations, our method complements existing content analysis techniques with a forward-looking dimension that captures what readers anticipate, not just what they have consumed. We validate the approach using both survey-elicited beliefs and a rational expectations framework, finding that LLM-derived expectations correlate with human beliefs and are informative predictors of realized story content across multiple narrative dimensions. Applying the framework to both survey and observational data, we find that forward-looking expectations are associated with reader engagement beyond what actual content features alone explain, suggesting that a richer understanding of engagement requires modeling what readers expect. 

We hope this framework serves as a starting point for empirical researchers interested in understanding how consumer beliefs about unstructured narrative content shape behavior, with applications spanning content creation, platform strategy, and the broader study of narrative media.

\section*{Funding and Competing Interests}
All authors certify that they have no affiliations with or involvement in any organization or entity with any financial interest or non-financial interest in the subject matter or materials discussed in this manuscript. The authors have no funding to report.

%% file: Sections/appendix_story_generation.tex
The goal of the story stimuli generation procedure is to produce stimuli that are representative across genres and exhibit sufficient variation in valence and arousal for our validation tests. We generate the stimuli in three stages.

\paragraph{Stage 1: Premise and first-chapter generation.} We generate 500 one-sentence story premises (100 per genre: fantasy, literary fiction, romance, science fiction, thriller) using \texttt{gpt-5.4-mini-2026-03-17} (temperature 1) with the system prompt \textit{``You are a fiction author writing short three-chapter stories.''} For each premise, we generate a first chapter of approximately 250 words using the same model: \textit{``Write Chapter 1 of a three-chapter story. Each chapter should be around 250 words. Chapter 1 should establish the characters, setting, and central conflict. Premise: [premise text]. Write Chapter 1 now.''}

\paragraph{Stage 2: Premise selection.} We select 40 premises (8 per genre) from the 500 candidates to maximize diversity in expected valence and arousal. For each candidate, we generate 50 imagined continuations using \texttt{gpt-4.1-mini-2025\\-04-14} (temperature 1) given its lower cost and extract valence and arousal from each continuation (temperature 0). We compute expected valence and arousal as the average across continuations and select premises using inverse-density sampling in this two-dimensional space.

\paragraph{Stage 3: Chapters 2 and 3 generation with emotional conditioning.} For each selected premise, we generate four second- and third-chapter variants using GPT-4.1-mini (temperature = 1.0). The four variants correspond to a $2 \times 2$ factorial design crossing valence (positive vs.\ negative) and arousal (high vs.\ low energy). We append the following emotional tone instructions to the generation prompt. The valence instructions read: \textit{``Words can be positive or negative. For example, very positive words include `vacation' and `happiness,' while very negative words include `torture' and `mourn.' Write in a [positive/negative] way.''} The arousal instructions read: \textit{``Words can be low or high energy. For example, very high energy words include `scare' and `exciting,' while very low energy words include `calm' and `dull.' Write in a [high-energy/low-energy] way.''} The full user prompt for chapters 2 and 3 is: \textit{``Here is Chapter 1 of a story: [Ch1 text]. Write Chapter 2 and Chapter 3, continuing from Chapter 1. Each chapter should be around 250 words. Chapter 3 should provide a resolution to the story's central conflict. [tone instructions]. Write the chapters now.''}

This yields 160 unique story variants (40 premises $\times$ 4 conditions) and 200 unique chapters (40 shared first chapters plus 160 second chapters).

%% file: AppendixSections/sensitivity_N.tex
This section presents the sensitivity analysis for the number of imagined continuations $N$. In our main analysis, we set $N = 50$ imagined continuations per chapter. To ensure that our results are robust to this choice, we examine how expected valence and arousal and the semantic path features stabilize as $N$ increases.

\subsection{Bootstrap Procedure}

To determine the number of imagined continuations needed, we conduct a bootstrap resampling analysis on a pilot sample of approximately 367 chapters from the observational data. We examine the emotion and semantic path features.
We assume 100 imagined continuations is sufficient to recover the ground truth and compute the ground-truth estimates using all 100 available imagined continuations for each chapter $c$: $
\text{Expectation}_c = \frac{1}{K} \sum_{k=1}^{K} z_{c}^{k}$, 
where $z_{c}^{k}$ is the feature value extracted from the $k$th imagined continuation of chapter $c$. We then simulate the scenario of having only $N$ imagined continuations. For each $N \in \{1, 2, \ldots, 100\}$, we run $B = 1{,}000$ bootstrap iterations with replacement.
Across the $B$ iterations, we compute the bias (average deviation from ground truth) and the bootstrap standard deviation (estimation noise) for the expectation estimates.

\subsection{Reliability Ratio}

To formalize the precision of the expectation estimate at a given $N$, we compute a reliability ratio that quantifies the fraction of observed chapter-to-chapter variance in the expectation that is true signal versus sampling noise. Let $\sigma^2_B$ denote the between-chapter variance of the true chapter-level expectations (i.e., the signal), and let $\bar{\sigma}^2_W$ denote the average within-chapter variance across imagined continuations (i.e., the noise source). When using only $N$ imagined continuations per chapter, the noise variance injected into each chapter's expectation estimate is $\bar{\sigma}^2_W / N$, and the reliability ratio is $\rho(N) = \frac{\sigma^2_B}{\sigma^2_B + \bar{\sigma}^2_W / N}$. A reliability of $\rho = 0.9$ indicates that 90\% of the observed chapter-to-chapter variance in the expectation estimate is true signal. By the classical measurement error attenuation result, OLS regression coefficients using this feature are attenuated by a factor of approximately $\rho$ (i.e., $\hat{\beta} \approx \rho \cdot \beta$). Higher reliability implies less bias in downstream regressions.

\subsection{Results}

At $N = 50$, all five features achieve a reliability ratio above 0.90, with valence reaching approximately 0.99 and circuitousness at approximately 0.90. The marginal improvement from $N = 50$ to $N = 100$ is negligible, adding only about 0.01 to the reliability across features. Both the bootstrap standard deviation and bias stabilize well before $N = 50$, indicating that additional imagined continuations beyond this point yield diminishing returns.

%% file: AppendixSections/survey_details.tex
This appendix provides details of the survey instrument used in the survey validation study. Participants were recruited to complete a study about story beliefs. After providing informed consent, participants read two short stories, each presented over multiple chapters. For each story, participants first read Chapter 1 and then answered questions about the emotional content of the chapter, their interest in continuing, and their beliefs about the emotional content of the next chapter. Participants then read Chapter 2 and answered analogous questions.
Finally, participants were given the option to read Chapter 3, which provides a revealed-preference measure of continued engagement.

\paragraph{Realized Chapter Ratings.} After each chapter, participants were asked to rate its valence and arousal. For valence, participants saw the following explanation: ``Words can be characterized as being different levels of negative or positive. For example, very negative words include ``torture'' and ``mourn,'' neutral words include ``vertical'' and ``beaver,'' and very positive words include ``vacation'' and ``happiness.'''' They were then asked: ``Where on the very negative to very positive spectrum do you think the words in the chapter are?'' Responses were given on a 1--5 scale.

For arousal, participants saw the following explanation: ``In addition to being negative or positive, words can also be characterized as being low or high energy. For example, very low energy words include ``calm'' and ``dull,'' medium energy words include ``relationship'' and ``religion,'' while very high energy words include ``scare'' and ``exciting.'''' They were then asked: ``Where on the very low energy to very high energy spectrum do you think the words in the chapter are?'' Responses were given on a 1--5 scale.

\paragraph{Stated Interest.} After each of the first two chapters, participants were asked: ``How interested are you in continuing to read the next chapter?'' Responses were given on a 1--5 scale.

\paragraph{Beliefs About Next-Chapter Valence and Arousal.} Participants were asked to assign probabilities to possible valence levels of the next chapter: ``Based on the chapter you just read, what do you think is the likelihood that the text of the next chapter will be very negative or very positive? Please assign a percentage to each. The total must sum to 100\%. '' They then assigned percentages across the following categories: very negative, somewhat negative, neutral, somewhat positive, and very positive.

Participants were also asked to assign probabilities to possible arousal levels of the next chapter: ``Based on the chapter you just read, what do you think is the likelihood that the text of the next chapter will be very low energy or very high energy? Please assign a percentage to each. The total must sum to 100\%.'' They then assigned percentages across the following categories: very low energy, somewhat low energy, medium energy, somewhat high energy, and very high energy.

\paragraph{Revealed Preference for Continuing.} After completing the Chapter 2 questions, participants were told that reading the final chapter was optional: ``The next chapter is the final chapter of this story and reading it is optional for this survey. If you decide to read the final chapter, it will take 1--2 minutes and we will ask you one additional question. If you decide not to read the final chapter, you will continue to the next story.'' Participants were then asked: ``Would you like to read the next chapter?'' The response options were ``No'' and ``Yes.''

%% file: AppendixSections/survey_validation_robustness.tex
\begin{figure}[h!]
    \centering
    \caption{LLM Expectations and Survey Expectations Across Genres}
    \label{fig:claim1_scatter_genre}
    \includegraphics[width=\textwidth]{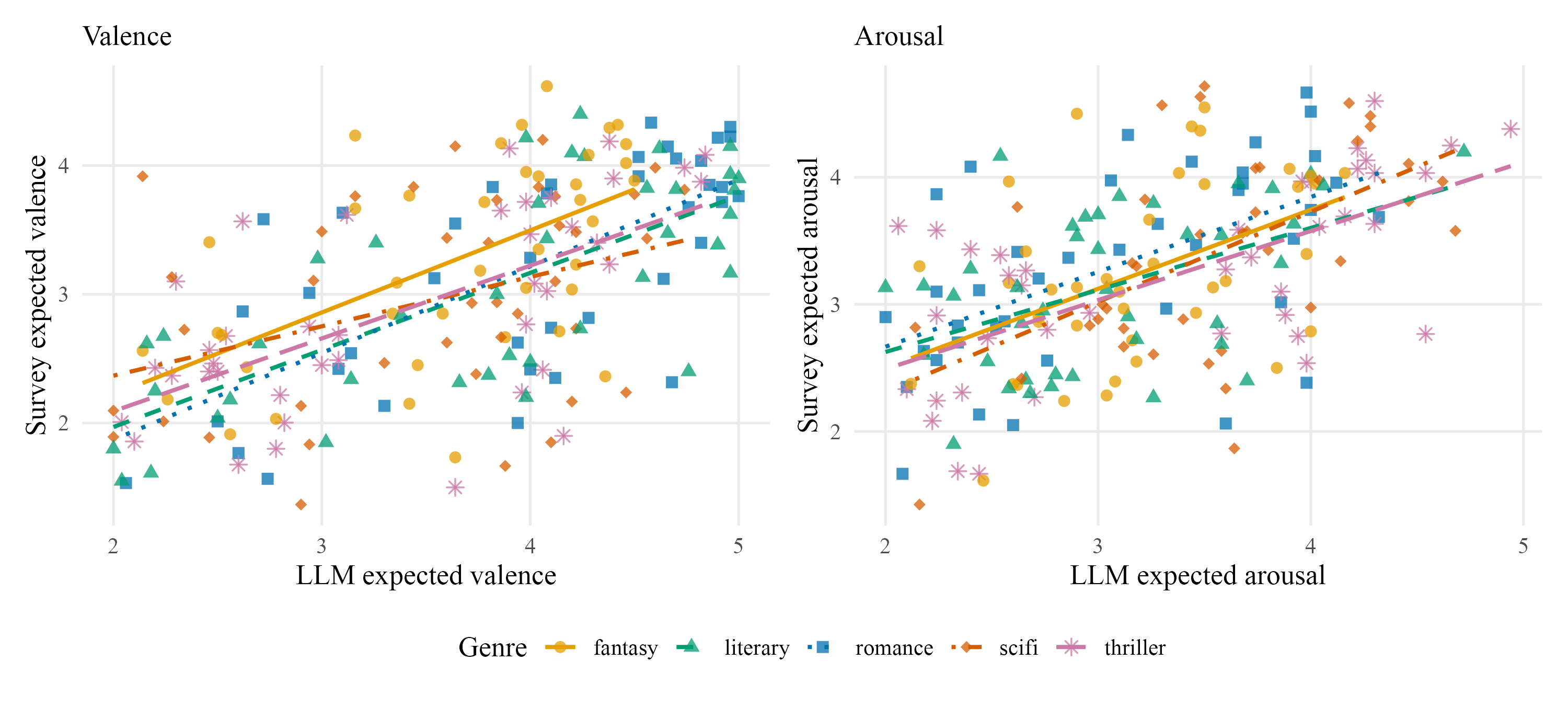}
\end{figure}

We conduct a number of robustness checks for the main findings obtained from the survey data. First, we show that the positive relationship between stated expectations and LLM-derived expectations hold across all five genres captured in the survey data in Figure \ref{fig:claim1_scatter_genre} and Table \ref{tab:claim1_combined}. 

\begin{table}[h!] \centering
\small
\begin{threeparttable}
  \caption{LLM-Derived Expectations Approximate Survey-Elicited Expectations Across Genres}
  \label{tab:claim1_combined}
\input{Tables/survey/claim1_combined_optA}
  \begin{tablenotes}[flushleft]
  \footnotesize
  \item \textit{Note:} $^{***}p<0.01$; $^{**}p<0.05$; $^{*}p<0.1$.
  \end{tablenotes}
\end{threeparttable}
\end{table}

In Table \ref{tab:claim1_validation_controls}, we add controls for LLM-rated current valence and arousal. Adding these variables reduces the slopes (from 0.57 to 0.26 for valence; from 0.56 to 0.32 for arousal), consistent with the expectations being partly informed by the current chapter's emotional content. Crucially, both slopes remain strongly significant ($p < 0.001$), confirming that the LLM expectations capture forward-looking beliefs beyond what current chapter features explain.

\begin{table}[htbp!] \centering
\small
\begin{threeparttable}
  \caption{Validation Survives Controlling for Current Valence and Arousal}
  \label{tab:claim1_validation_controls}
\input{Tables/survey/claim1_validation_controls}
  \begin{tablenotes}[flushleft]
  \footnotesize
  \item \textit{Note:} $^{***}p<0.01$; $^{**}p<0.05$; $^{*}p<0.1$.
  \end{tablenotes}
\end{threeparttable}
\end{table}

Consistent with the literature that has found that off-the-shelf LLMs tend to produce overly concentrated distributions relative to humans \citep{bisbee2024synthetic, deng2026examining}, we find that the LLM-derived uncertainty measures for valence and arousal have lower variance than the equivalent based on human response. To further assess whether LLM-derived uncertainty can proxy human uncertainty, we regress survey uncertainty on LLM-derived uncertainty for valence and arousal. Table \ref{tab:uncertainty_validation} presents the regression coefficients. We find that the coefficients are small in magnitude, statistically insignificant, and explain little of the variation in survey responses. Therefore, the distribution of imagined continuations we generated cannot approximate the human stated uncertainty. 

\begin{table}[!htbp] \centering
\caption{LLM-derived Uncertainty and Survey-derived Uncertainty}
\label{tab:uncertainty_validation}
\small
\input{Tables/Appendix/10e_survey_validation_uncertainty}
\end{table}

%% file: Tables/survey/claim1_combined_optA.tex
\begin{tabular}{lccccc}
\toprule
 & \multicolumn{2}{c}{Expected Valence} & \multicolumn{2}{c}{Expected Arousal} & \\
\cmidrule(lr){2-3} \cmidrule(lr){4-5}
 & $\hat{\beta}$ & $R^2$ & $\hat{\beta}$ & $R^2$ & $N$ \\
\midrule
\textit{All genres} & 0.571$^{***}$ & 0.369 & 0.563$^{***}$ & 0.297 & 200 \\
 & & & & & \\[-6pt]
\quad fantasy & 0.638$^{***}$ & 0.321 & 0.622$^{**}$ & 0.214 & 40 \\
\quad literary & 0.598$^{***}$ & 0.522 & 0.488$^{**}$ & 0.228 & 40 \\
\quad romance & 0.673$^{***}$ & 0.476 & 0.589$^{***}$ & 0.311 & 40 \\
\quad scifi & 0.384$^{*}\ \ $ & 0.143 & 0.708$^{***}$ & 0.300 & 40 \\
\quad thriller & 0.565$^{***}$ & 0.403 & 0.543$^{***}$ & 0.429 & 40 \\
\bottomrule
\end{tabular}

%% file: Tables/survey/claim1_validation_controls.tex
\begin{tabular}{lcccc}
\hline \hline
 & \multicolumn{2}{c}{Survey expected valence} & \multicolumn{2}{c}{Survey expected arousal} \\
\cmidrule(lr){2-3} \cmidrule(lr){4-5}
 & Baseline & + Controls & Baseline & + Controls \\
 & (1) & (2) & (3) & (4) \\
\hline
LLM expected valence & 0.571*** & 0.255*** &  &  \\
 & (0.053) & (0.039) &  &  \\
LLM expected arousal &  &  & 0.563*** & 0.322*** \\
 &  &  & (0.062) & (0.052) \\
Current valence &  & 0.371*** &  & 0.090*** \\
 &  & (0.022) &  & (0.022) \\
Current arousal &  & 0.053** &  & 0.315*** \\
 &  & (0.021) &  & (0.027) \\
\hline
Observations & 200 & 200 & 200 & 200 \\
R$^2$ & 0.369 & 0.755 & 0.297 & 0.619 \\
Adjusted R$^2$ & 0.366 & 0.752 & 0.293 & 0.613 \\
\hline \hline
\end{tabular}

%% file: Tables/Appendix/10e_survey_validation_uncertainty.tex
\begin{tabular}{@{\extracolsep{5pt}}lcc} 
\hline 
\hline
 & \multicolumn{2}{c}{\textit{Dependent variable:}} \\ 
\cline{2-3} 
& Survey uncertainty valence & Survey uncertainty arousal \\ 
& (1) & (2)\\ 
\hline
 LLM uncertainty valence & 0.142 &  \\ 
  & (0.146) &  \\ 
  LLM uncertainty arousal &  & 0.077 \\ 
  &  & (0.259) \\ 
  Constant & 0.799$^{***}$ & 0.664$^{***}$ \\ 
  & (0.051) & (0.056) \\ 
 \hline
Observations & 200 & 200 \\ 
R$^{2}$ & 0.005 & 0.0004 \\ 
Adjusted R$^{2}$ & $-$0.0003 & $-$0.005 \\ 
\hline 
\hline
\multicolumn{3}{l}{\textit{Note:} $^{***}p<0.01$; $^{**}p<0.05$; $^{*}p<0.1$.} \\
\end{tabular} 

%% file: AppendixSections/feature_extraction.tex
This appendix provides implementation details for the imagination generation and feature extraction procedures described in Section~\ref{section:method}.

\subsection{Imagination Generation}

We generate imagined story continuations using \texttt{gpt-4o-mini-2024-07-18}. For each chapter $t$ in book $i$, we provide the full text of chapters $1$ through $t$ as context and prompt the model to generate the next chapter using the prompt described in Section~\ref{section:method}.

We generate $N = 50$ continuations per chapter using OpenAI's \texttt{n} parameter, which produces all continuations in a single API call with identical context. The generation parameters are: temperature $= 1.0$, top\_p $= 1.0$, frequency penalty $= 0.0$, and presence penalty $= 0.0$. Each continuation can be up to 16{,}000 tokens. The context window accommodates up to 90{,}000 words of preceding chapter text; when the preceding text exceeds this limit, we truncate from the beginning, starting from the first complete sentence after the truncation point.

\subsection{Emotion Features: Valence and Arousal}\label{appendix:va_extraction}

We extract valence and arousal from each chapter and imagined continuation using \texttt{gpt-4.1-mini-2025-04-14} (temperature $0$). The model receives the chapter text and returns integer ratings for each dimension on a 1--5 scale. The prompt is as follows:

\begin{quote}
\textbf{System prompt:} ``Rate the overall emotional tone of the following text on two dimensions:

1. Valence (1--5): how negative vs positive (1 = very negative, 5 = very positive)

2. Arousal (1--5): how calm vs intense (1 = very calm, 5 = very intense)

Respond ONLY with a JSON object: \{``valence'': $\langle$int$\rangle$, ``arousal'': $\langle$int$\rangle$\}''

\textbf{User prompt:} [chapter text]
\end{quote}

\noindent The model returns a JSON object with two integer fields. Valence captures how negative (1) versus positive (5) the overall emotional tone of the chapter is. Arousal captures how calm (1) versus intense (5) the emotional energy is. We apply this extraction to both the actual chapter text and each of the $N$ imagined continuations independently.

\subsection{Semantic Path Features}\label{appendix:semantic_path_extraction}

We measure the speed, volume, and circuitousness of each chapter and imagined continuation following the approach proposed by \citet{toubia2021quantifying}. These measures capture the dynamics of narrative content within a chapter by analyzing the trajectory of text embeddings.

To operationalize, we embed story text using \texttt{text-embedding-3-large}, which produces 3{,}072-dimensional embedding vectors. We pass the original chapter text as written, truncating any segment that exceeds 8{,}000 tokens to stay within the model's 8{,}192-token context limit. Each chapter is divided into $T = 20$ equal-sized segments by splitting its text into $T$ consecutive spans of equal word count. We embed each chunk's text directly, obtaining one 3{,}072-dimensional vector per chunk. We then use these embeddings to compute speed, volume, and circuitousness. 

\subsection{11 Narrative Features}\label{appendix:hyp_gen_extraction}

We extract eleven narrative dimensions (narrative tension, curiosity, emotional complexity, character agency, social conflict, resolution, cliffhanger intensity, narrative vividness, pacing, character relatability, cognitive complexity) using \texttt{gpt-5.4-nano-2026-03-17} (temperature $0$). The dimensions are defined based on the following text:

\textbf{Narrative tension / Stakes (1--5):} 1 = nothing at stake, no consequences, 2 = low stakes, minor concerns, 3 = moderate stakes, 4 = high stakes, significant consequences for characters, 5 = extreme stakes, life-changing or life-threatening.

\textbf{Curiosity / Information gaps (1--5):} 1 = everything is explained, no unanswered questions, 2 = minor open questions, 3 = some curiosity-provoking elements, 4 = significant unanswered questions or mysteries, 5 = major information gaps that demand resolution.

\textbf{Emotional complexity (1--5):} 1 = simple, single emotion dominates clearly, 2 = mostly one emotion with slight nuance, 3 = mixed emotions, some tension between feelings, 4 = complex emotional layering, conflicting feelings, 5 = deeply complex, multiple conflicting emotions intertwined.

\textbf{Character agency (1--5):} 1 = characters are entirely passive, things happen to them, 2 = mostly reactive, limited initiative, 3 = moderate agency, some active choices, 4 = characters actively drive the plot through their decisions, 5 = strong agency, characters' choices are the primary plot engine.

\textbf{Social conflict (1--5):} 1 = no interpersonal tension, harmonious relationships, 2 = minor disagreements or mild tension, 3 = moderate interpersonal conflict, 4 = significant relational tension, confrontation, or betrayal, 5 = intense social conflict, deep interpersonal rupture.

\textbf{Resolution vs. Complication (1--5):} 1 = fully resolves existing plot threads, provides closure, 2 = mostly resolves with minor new elements, 3 = balanced --- resolves some threads while introducing others, 4 = mostly introduces new complications or questions, 5 = purely complicating --- opens new threads without resolving any.

\textbf{Cliffhanger intensity (1--5):} 1 = fully resolved ending, no open questions, 2 = mostly resolved, minor loose ends, 3 = some unresolved tension or open questions, 4 = significant unresolved tension or a question that demands the next chapter, 5 = major cliffhanger --- the chapter ends at a critical moment of suspense.

\textbf{Narrative vividness (1--5):} 1 = abstract, tell-not-show, minimal sensory detail, 2 = some concrete details but mostly summary, 3 = moderate sensory detail and scene-setting, 4 = vivid and immersive, strong sensory language, 5 = extremely vivid, deeply immersive, rich sensory and emotional detail.

\textbf{Pacing (1--5):} 1 = very slow, contemplative, mostly internal reflection, 2 = slow, unhurried, deliberate unfolding, 3 = moderate pace, balanced action and reflection, 4 = fast-paced, events progress quickly, 5 = very fast, rapid-fire events, high urgency.

\textbf{Character relatability (1--5):} 1 = characters feel distant, generic, or hard to empathize with, 2 = somewhat flat characters with limited emotional depth, 3 = moderately relatable, some emotional resonance, 4 = relatable characters with clear motivations and emotional depth, 5 = deeply relatable, strong emotional connection, compelling inner life.

\textbf{Cognitive complexity (1--5):} 1 = simple, straightforward, no ambiguity, 2 = mostly straightforward with minor complexity, 3 = moderate complexity, some ambiguity or layering, 4 = complex, requires attention, plot twists or moral ambiguity, 5 = very complex, multiple layers of meaning, significant ambiguity.

%% file: AppendixSections/extraction_narrative.tex
\subsection{Extraction of 11 Narrative Dimensions}

We extract eleven narrative dimensions (narrative tension, curiosity, emotional complexity, character agency, social conflict, resolution, cliffhanger intensity, narrative vividness, pacing, character relatability, cognitive complexity) using \texttt{gpt-5.4-nano-2026-03-17} (temperature $0$). The prompts for the generative and direct-elicitation implementations share the same system-prompt structure and differ only in the task they pose and the response format they request.

\textbf{Generative implementation.} The prompt asks the model to \emph{rate} the most recent chapter it has been shown. That chapter is either the actual chapter or one imagined continuation, presented in an identical format so that the model cannot tell the two apart. For each dimension, it returns a single integer rating on a 1--5 scale. Applying this prompt to the imagined continuations of a chapter and averaging the ratings yields the expected feature value, while applying it to the actual chapter yields the realized value.

\textbf{Direct elicitation.} The prompt asks the model to \emph{predict} the next, unread chapter from the chapters read so far. For each dimension, it directly returns a five-bin probability distribution (five percentages, one per rating level, summing to 100).

The system prompt has three components: a general instruction (which differs by implementation), the feature definitions (shared across both implementations), and an output specification (which differs by implementation). We first give the shared feature description, then the complete system prompt and user prompt for each implementation.

\subsubsection{Shared Feature Description}\label{appendix:shared_feature_def}

Both variants use the same feature definitions (Appendix \ref{appendix:hyp_gen_extraction}). 

\subsubsection{Generative Implementation and Actual Content}

This variant rates a single chapter, either the actual chapter $t$ or one imagined continuation of it, on each of the eleven dimensions, given the preceding chapters as context.

\textbf{System prompt.} The system prompt concatenates the general instruction, the shared feature description, and the output specification:

\begin{quote}
You are an average reader. You have been reading a story chapter by chapter. The chapters you have read so far include those in \texttt{<preceding\_chapters\_as\_context>} and \texttt{<latest\_chapter\_to\_rate>}, and the chapter to rate appears in \texttt{<latest\_chapter\_to\_rate>}. What is your rating for each of the eleven dimensions defined below for the chapter in \texttt{<latest\_chapter\_to\_rate>}?

\begin{quote}
\textit{[shared feature description]}
\end{quote}

Provide your ratings for the chapter in \texttt{<latest\_chapter\_to\_rate>}. For each dimension, return a single integer between 1 (lowest) and 5 (highest) that represents your rating for the chapter in \texttt{<latest\_chapter\_to\_rate>}. Respond ONLY with a JSON object. Each key maps to a single integer: \texttt{\{"tension": <int>, "curiosity": <int>, \ldots, "cog\_complexity": <int>\}}.
\end{quote}

\textbf{User prompt.}

\begin{quote}
\texttt{<preceding\_chapters\_as\_context>} \\
\textit{[text of chapters 1 through $t-1$]} \\
\texttt{</preceding\_chapters\_as\_context>} \\
Rate the latest chapter: \\
\texttt{<latest\_chapter\_to\_rate>} \\
\textit{[text of the actual chapter $t$ or one imagined continuation]} \\
\texttt{</latest\_chapter\_to\_rate>}
\end{quote}

\subsubsection{Direct Elicitation}

This variant predicts the next, unread chapter, returning for each of the eleven dimensions a probability distribution over the 1--5 rating scale.

\textbf{System prompt.} The system prompt concatenates the general instruction, the shared feature description, and the output specification:

\begin{quote}
You are an average reader. You have been reading a story chapter by chapter. The chapters you have read so far appear in the user message. What is your belief for each of the eleven dimensions defined below for the NEXT chapter?

\begin{quote}
\textit{[shared feature description]}
\end{quote}

Provide your beliefs for the NEXT chapter, which you have not yet read. For each dimension, return five integers that represent the chance of each rating happening in the next chapter --- these are PERCENTAGES, not ratings, that must sum to 100. The five elements correspond to rating 1 (first element) through rating 5 (last element). Example for one dimension: [10, 30, 30, 20, 10] means 10\% chance the next chapter rates 1, 30\% chance it rates 2, 30\% chance it rates 3, 20\% chance it rates 4, 10\% chance it rates 5. The five percentages must sum to 100 (10 + 30 + 30 + 20 + 10 = 100). Respond ONLY with a JSON object. Each key maps to an array of five percentages summing to 100, ordered from rating 1 to rating 5: \texttt{\{"tension": [p1, p2, p3, p4, p5], \ldots, "cog\_complexity": [p1, p2, p3, p4, p5]\}}.
\end{quote}

\textbf{User prompt.}

\begin{quote}
\texttt{<story\_so\_far>} \\
\textit{[text of chapters 1 through $t$]} \\
\texttt{</story\_so\_far>}
\end{quote}

%% file: AppendixSections/data_preprocessing.tex
We preprocess the observational data at the text, chapter, and book levels. At the text level, we remove non-story material such as trailing author notes, HTML tags, HTML entities, and emojis. At the chapter level, we distinguish plot-related story chapters from contextual material, such as prologues, character lists, author notes, and other non-plot text. Contextual chapters are retained when useful for generating imagined continuations, but are excluded from feature extraction and analysis. We also exclude chapters that are too short, extreme length outliers, non-English, explicit, incoherent as narrative text, associated with zero reads, or linked to invalid engagement records. To avoid overlap with the model's training data, we exclude chapters created before December 1, 2023.

At the book level, we remove books that are marked as mature or contain explicit-content keywords in their metadata. We also require books to have sufficient readership (first-chapter read count $>$ 50), have at least 10 plot-related story chapters after filtering, and plausible chapter length. For analysis, we retain the first 9 plot-related chapters of each eligible book. We further exclude books that are predominantly non-English or have average chapter lengths that are unusually short or long, which often indicate non-narrative content or entire books uploaded as single chapters.

%% file: AppendixSections/semantic_sensitivity.tex
\begin{figure}[h!]
  \centering
  \caption{Semantic Speed and Circuitousness Coefficient Sensitivity} 
  \includegraphics[width=.75\textwidth]{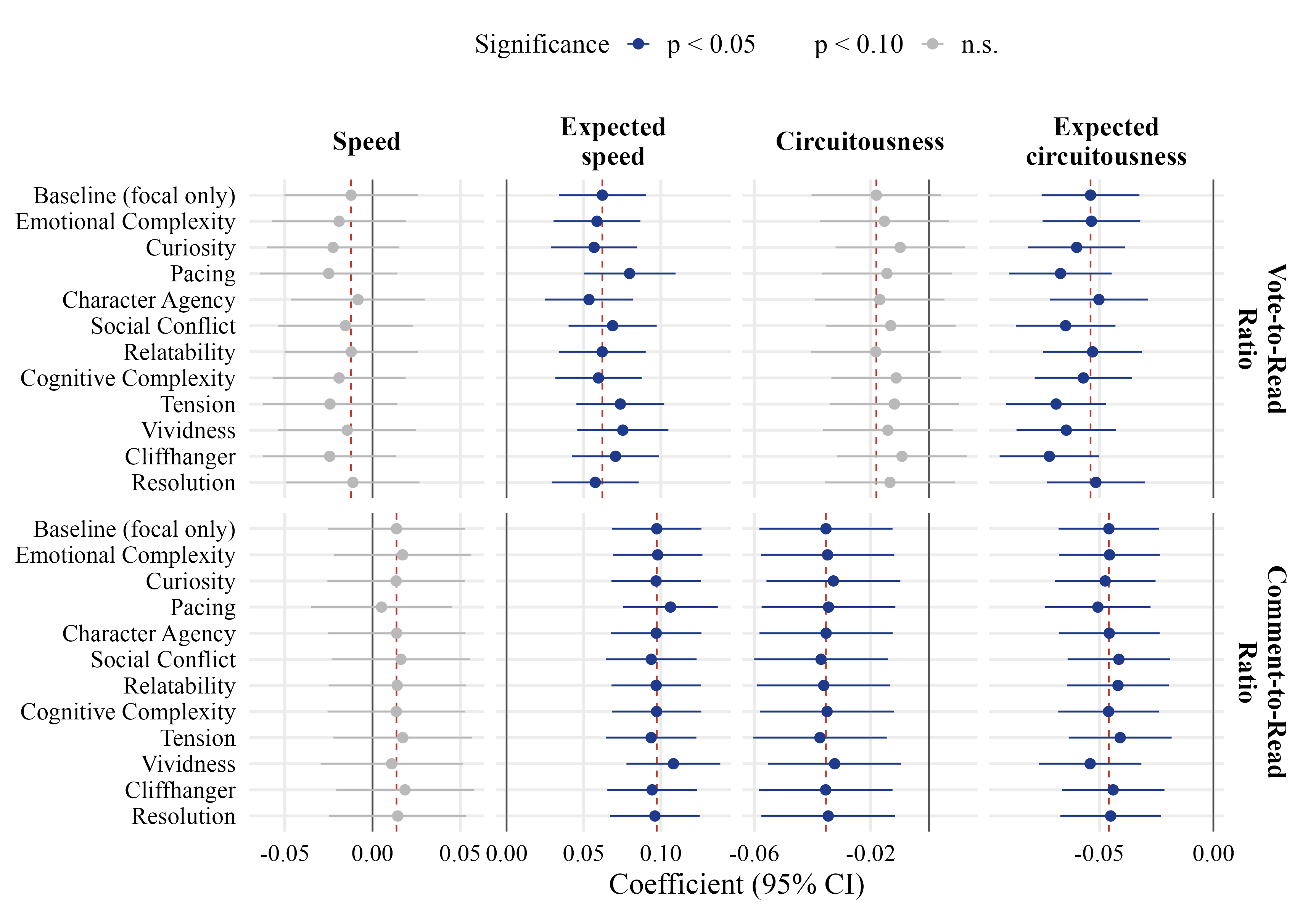}
  \vspace{0.5em}
    \begin{minipage}{\textwidth}
    \footnotesize
    \textit{Note}: Sensitivity of the semantic path coefficients to adding one narrative dimension at a time. The baseline specification regresses each engagement ratio on current
    and expected speed and circuitousness, with word count and log readership controls and chapter fixed effects.
    Each row adds the realized and expected ratings of a single dimension to that
    baseline. The dot is the partial coefficient, the whisker its 95\% confidence
    interval (two-way clustered by user and chapter), and the dashed red line highlights the baseline
    estimate. $N=8{,}399$.
    \end{minipage}
    \label{fig:obs_semantic_sensitivity}
\end{figure}

%% file: AppendixSections/embedding_pca.tex
We embed each actual chapter and each imagined continuation with \texttt{text-embedding-3-large}, which maps a passage of text to a 3{,}072-dimensional vector. Because this dimensionality is far larger than the number of stories in either empirical setting, we do not use the raw embeddings as regressors; instead we reduce them with principal component analysis (PCA). We apply PCA separately within each empirical setting, the survey data and the observational data, and in each case we fit the PCA on the embeddings of the \emph{actual} chapter text only since the actual text should capture the variation in dimensions that stories differ over. 
To choose how many components to retain, we inspect the scree plot of eigenvalues and retain the number of dimensions up till elbow - 1. The elbow plot suggests retaining five components for the survey data and three components for the observational data.

\begin{table}[h!]
\centering
\small
\caption{Which Embedding Components Are Associated with Survey Stated Interest?}
\label{tab:embedding_interest_obs_pca}
\input{Tables/Alternative/embedding_interest_apca_k5_wide}
\end{table}

\begin{table}[h!] \centering
\small
\caption{Which Embedding Components Are Associated with Voting and Commenting?}
\label{tab:09-engagement}
\input{Tables/Alternative/09_engagement_regression_fragment_wide}
\end{table}

%% file: Tables/Alternative/embedding_interest_apca_k5_wide.tex
\begin{tabular}{l @{\hspace{8pt}} cc @{\hspace{8pt}} cc}
\hline \hline
 & \multicolumn{4}{c}{DV: Stated Interest} \\
\cmidrule(lr){2-5}
 & \multicolumn{2}{c}{(1) Current PCs} & \multicolumn{2}{c}{(2) $+$ Expectations} \\
\cmidrule(lr){2-3} \cmidrule(lr){4-5}
 & Coef. & SE & Coef. & SE \\
\hline
PC 1          & $-1.292^{**}$ & (0.562) & $-1.454^{**}$ & (0.590) \\
PC 2          & $0.336$       & (0.485) & $0.355$       & (0.640) \\
PC 3          & $-0.670^{**}$ & (0.270) & $-0.591$      & (0.393) \\
PC 4          & $0.536$       & (0.396) & $-0.092$      & (0.606) \\
PC 5          & $0.903^{**}$  & (0.446) & $1.194$       & (0.751) \\
Expected PC 1 &               &         & $0.054$       & (1.021) \\
Expected PC 2 &               &         & $0.028$       & (0.897) \\
Expected PC 3 &               &         & $-0.570$      & (1.141) \\
Expected PC 4 &               &         & $0.951$       & (0.848) \\
Expected PC 5 &               &         & $-0.328$      & (0.908) \\
\hline
Chapter number FE & \multicolumn{2}{c}{Yes}   & \multicolumn{2}{c}{Yes}   \\
User FE           & \multicolumn{2}{c}{Yes}   & \multicolumn{2}{c}{Yes}   \\
Observations      & \multicolumn{2}{c}{1,052} & \multicolumn{2}{c}{1,052} \\
R$^{2}$           & \multicolumn{2}{c}{0.723} & \multicolumn{2}{c}{0.724} \\
Adjusted R$^{2}$  & \multicolumn{2}{c}{0.628} & \multicolumn{2}{c}{0.627} \\
\hline \hline
\multicolumn{5}{l}{\textit{Note:} Two-way clustered standard errors for user and} \\
\multicolumn{5}{l}{chapter. $^{***}p<0.01$; $^{**}p<0.05$; $^{*}p<0.1$.} \\
\end{tabular}

%% file: Tables/Alternative/09_engagement_regression_fragment_wide.tex
\begin{tabular}{@{\extracolsep{3pt}}l cc cc cc cc}
\hline \hline
 & \multicolumn{4}{c}{Vote-to-Read} & \multicolumn{4}{c}{Comment-to-Read} \\
\cmidrule(lr){2-5} \cmidrule(lr){6-9}
 & \multicolumn{2}{c}{Baseline} & \multicolumn{2}{c}{$+$ Expectations} & \multicolumn{2}{c}{Baseline} & \multicolumn{2}{c}{$+$ Expectations} \\
\cmidrule(lr){2-3} \cmidrule(lr){4-5} \cmidrule(lr){6-7} \cmidrule(lr){8-9}
 & Coef. & SE & Coef. & SE & Coef. & SE & Coef. & SE \\
\hline
\multicolumn{9}{l}{\textit{Embedding Principal Components}} \\
PC 1          & $-0.117^{***}$ & $(0.011)$ & $-0.112^{***}$ & $(0.029)$ & $-0.091^{***}$ & $(0.012)$ & $-0.038$       & $(0.030)$ \\
PC 2          & $0.075^{***}$  & $(0.010)$ & $0.207^{***}$  & $(0.035)$ & $-0.043^{***}$ & $(0.011)$ & $0.026$        & $(0.037)$ \\
PC 3          & $-0.121^{***}$ & $(0.012)$ & $-0.053^{***}$ & $(0.019)$ & $-0.047^{***}$ & $(0.013)$ & $-0.029$       & $(0.020)$ \\
Expected PC 1 &                &           & $-0.108^{***}$ & $(0.031)$ &                &           & $-0.107^{***}$ & $(0.032)$ \\
Expected PC 2 &                &           & $-0.165^{***}$ & $(0.036)$ &                &           & $-0.089^{**}$  & $(0.037)$ \\
Expected PC 3 &                &           & $-0.148^{***}$ & $(0.020)$ &                &           & $-0.069^{***}$ & $(0.021)$ \\[4pt]
\hline
Chapter Number FE & \multicolumn{2}{c}{Yes}    & \multicolumn{2}{c}{Yes}    & \multicolumn{2}{c}{Yes}    & \multicolumn{2}{c}{Yes}    \\
Observations      & \multicolumn{2}{c}{8,399}  & \multicolumn{2}{c}{8,399}  & \multicolumn{2}{c}{8,399}  & \multicolumn{2}{c}{8,399}  \\
R$^{2}$           & \multicolumn{2}{c}{0.1293} & \multicolumn{2}{c}{0.1371} & \multicolumn{2}{c}{0.0478} & \multicolumn{2}{c}{0.0501} \\
Adjusted R$^{2}$  & \multicolumn{2}{c}{0.1279} & \multicolumn{2}{c}{0.1355} & \multicolumn{2}{c}{0.0463} & \multicolumn{2}{c}{0.0483} \\
$F$ (vs baseline) &               &            & \multicolumn{2}{c}{$25.30^{***}$} &              &           & \multicolumn{2}{c}{$6.82^{***}$} \\
\hline \hline
\multicolumn{9}{l}{\textit{Note:} $^{***}p<0.01$; $^{**}p<0.05$; $^{*}p<0.1$. Log first-chapter read count and word count controlled.} \\
\end{tabular}

%% file: reference.bib
@article{ely2015suspense,
  title={Suspense and surprise},
  author={Ely, Jeffrey and Frankel, Alexander and Kamenica, Emir},
  journal={Journal of Political Economy},
  volume={123},
  number={1},
  pages={215--260},
  year={2015},
  publisher={University of Chicago Press Chicago, IL}
}

@article{rust1987optimal,
  title={Optimal replacement of GMC bus engines: An empirical model of Harold Zurcher},
  author={Rust, John},
  journal={Econometrica: Journal of the Econometric Society},
  pages={999--1033},
  year={1987},
  publisher={JSTOR}
}

@article{hitsch2006empirical,
  title={An empirical model of optimal dynamic product launch and exit under demand uncertainty},
  author={Hitsch, G{\"u}nter J},
  journal={Marketing Science},
  volume={25},
  number={1},
  pages={25--50},
  year={2006},
  publisher={INFORMS}
}

@article{misra2011structural,
  title={A structural model of sales-force compensation dynamics: Estimation and field implementation},
  author={Misra, Sanjog and Nair, Harikesh S},
  journal={Quantitative Marketing and Economics},
  volume={9},
  pages={211--257},
  year={2011},
  publisher={Springer}
}

@article{erdem1996decision,
  title={Decision-making under uncertainty: Capturing dynamic brand choice processes in turbulent consumer goods markets},
  author={Erdem, T{\"u}lin and Keane, Michael P},
  journal={Marketing science},
  volume={15},
  number={1},
  pages={1--20},
  year={1996},
  publisher={INFORMS}
}

@article{lee2024generative,
  title={Generative Brand Choice},
  author={Lee, Kevin},
  year={2024}
}

@article{fan2018hierarchical,
  title={Hierarchical neural story generation},
  author={Fan, Angela and Lewis, Mike and Dauphin, Yann},
  journal={arXiv preprint arXiv:1805.04833},
  year={2018}
}

@article{russell1980circumplex,
  title={A circumplex model of affect.},
  author={Russell, James A},
  journal={Journal of personality and social psychology},
  volume={39},
  number={6},
  pages={1161},
  year={1980},
  publisher={American Psychological Association}
}

@article{peng2025mega,
  title={Digital Twins as Funhouse Mirrors: Five Key Distortions},
  author={Peng, Tianyi and Gui, George and Brucks, Melanie and Merlau, Daniel J and Jiarui Fan, Grace and Ben Sliman, Malek and Johnson, Eric J and Althenayyan, Abdullah and Bellezza, Silvia and others},
  journal={arXiv preprint arXiv:2509.19088},
  year={2025}
}

@incollection{friedman1957permanent,
  title={The permanent income hypothesis},
  author={Friedman, Milton},
  booktitle={A theory of the consumption function},
  pages={20--37},
  year={1957},
  publisher={Princeton University Press}
}

@article{berger2021makes,
  title={What makes content engaging? How emotional dynamics shape success},
  author={Berger, Jonah and Kim, Yoon Duk and Meyer, Robert},
  journal={Journal of Consumer Research},
  volume={48},
  number={2},
  pages={235--250},
  year={2021},
  publisher={Oxford University Press}
}

@article{halawi2024approaching,
  title={Approaching Human-Level Forecasting with Language Models},
  author={Halawi, Danny and Zhang, Fred and Yueh-Han, Chen and Steinhardt, Jacob},
  journal={arXiv preprint arXiv:2402.18563},
  year={2024}
}

@article{li2024frontiers,
  title={Frontiers: Determining the validity of large language models for automated perceptual analysis},
  author={Li, Peiyao and Castelo, Noah and Katona, Zsolt and Sarvary, Miklos},
  journal={Marketing Science},
  volume={43},
  number={2},
  pages={254--266},
  year={2024},
  publisher={INFORMS}
}

@book{waldfogel2018digital,
  title={Digital Renaissance: What Data and Economics Tell Us about the Future of Popular Culture},
  author={Waldfogel, Joel},
  year={2018},
  publisher={Princeton University Press}
}

@article{goldberg1987happy,
  title={Happy and sad TV programs: How they affect reactions to commercials},
  author={Goldberg, Marvin E and Gorn, Gerald J},
  journal={Journal of consumer research},
  volume={14},
  number={3},
  pages={387--403},
  year={1987},
  publisher={The University of Chicago Press}
}

@article{timoshenko2026transforming,
  title={Transforming the Voice of the Customer: Large Language Models for Identifying Customer Needs},
  author={Timoshenko, Artem and Mao, Chengfeng and Hauser, John R},
  year={2026}
}

@article{boughanmi2025reviews,
  title={From Reviews to Actionable Insights: An LLM-Based Approach for Attribute and Feature Extraction},
  author={Boughanmi, Khaled and Jedidi, Kamel and Jedidi, Nour},
  journal={arXiv preprint arXiv:2510.16551},
  year={2025}
}

@article{gao2025take,
  title={Take caution in using LLMs as human surrogates},
  author={Gao, Yuan and Lee, Dokyun and Burtch, Gordon and Fazelpour, Sina},
  journal={Proceedings of the National Academy of Sciences},
  volume={122},
  number={24},
  pages={e2501660122},
  year={2025},
  publisher={National Academy of Sciences}
}

@article{knight2025building,
  title={Building Persuasive Stories with Emotion Sequences},
  author={Knight, Samsun and Liu, Liu and Kornish, Laura J},
  journal={Available at SSRN 5489708},
  year={2025}
}

@article{gowrisankaran2012dynamics,
  title={Dynamics of consumer demand for new durable goods},
  author={Gowrisankaran, Gautam and Rysman, Marc},
  journal={Journal of Political Economy},
  volume={120},
  number={6},
  pages={1173--1219},
  year={2012},
  publisher={University of Chicago Press}
}

@article{hendel2006measuring,
  title={Measuring the implications of sales and consumer inventory behavior},
  author={Hendel, Igal and Nevo, Aviv},
  journal={Econometrica},
  volume={74},
  number={6},
  pages={1637--1673},
  year={2006},
  publisher={The Econometric Society}
}

@inproceedings{piper2021narrative,
  title={Narrative Theory for Computational Narrative Understanding},
  author={Piper, Andrew and So, Richard Jean and Bamman, David},
  booktitle={Proceedings of the 2021 Conference on Empirical Methods in Natural Language Processing (EMNLP)},
  pages={298--311},
  year={2021}
}

@article{eliashberg2007story,
  title={From Story Line to Box Office: A New Approach for Green-Lighting Movie Scripts},
  author={Eliashberg, Jehoshua and Hui, Sam K. and Zhang, Z. John},
  journal={Management Science},
  volume={53},
  number={6},
  pages={881--893},
  year={2007}
}

@article{shachar2025sell,
  title={Sell Me a Story},
  author={Shachar, Ron and Muchnik, Lev and Netzer, Oded},
  journal={SSRN Working Paper},
  number={5236601},
  year={2025}
}

@article{tan2008entertainment,
  title={Entertainment is Emotion: The Functional Architecture of the Entertainment Experience},
  author={Tan, Ed S.},
  journal={Media Psychology},
  volume={11},
  number={1},
  pages={28--51},
  year={2008}
}

@article{berger2012makes,
  title={What Makes Online Content Viral?},
  author={Berger, Jonah and Milkman, Katherine L.},
  journal={Journal of Marketing Research},
  volume={49},
  number={2},
  pages={192--205},
  year={2012}
}

@article{berger2023holds,
  title={What Holds Attention? Linguistic Drivers of Engagement},
  author={Berger, Jonah and Moe, Wendy W. and Schweidel, David A.},
  journal={Journal of Marketing},
  volume={87},
  number={5},
  pages={793--809},
  year={2023}
}

@article{knight2024narrative,
  title={Narrative Reversals and Story Success},
  author={Knight, Samsun and Rocklage, Matthew D. and Bart, Yakov},
  journal={Science Advances},
  volume={10},
  number={34},
  pages={eadl2013},
  year={2024}
}

@article{reagan2016emotional,
  title={The Emotional Arcs of Stories are Dominated by Six Basic Shapes},
  author={Reagan, Andrew J. and Mitchell, Lewis and Kiley, Dilan and Danforth, Christopher M. and Dodds, Peter Sheridan},
  journal={EPJ Data Science},
  volume={5},
  number={1},
  pages={31},
  year={2016}
}

@article{nabi2015role,
  title={The Role of a Narrative's Emotional Flow in Promoting Persuasive Outcomes},
  author={Nabi, Robin L. and Green, Melanie C.},
  journal={Media Psychology},
  volume={18},
  number={2},
  pages={137--162},
  year={2015}
}

@article{toubia2021quantifying,
  title={How Quantifying the Shape of Stories Predicts Their Success},
  author={Toubia, Olivier and Berger, Jonah and Eliashberg, Jehoshua},
  journal={Proceedings of the National Academy of Sciences},
  volume={118},
  number={26},
  pages={e2011695118},
  year={2021}
}

@article{toubia2019extracting,
  title={Extracting Features of Entertainment Products: A Guided Latent {D}irichlet Allocation Approach Informed by the Psychology of Media Consumption},
  author={Toubia, Olivier and Iyengar, Garud and Bunnell, Renee and Lemaire, Alain},
  journal={Journal of Marketing Research},
  volume={56},
  number={1},
  pages={18--36},
  year={2019}
}

@inproceedings{bamman2014bayesian,
  title={A {B}ayesian Mixed Effects Model of Literary Character},
  author={Bamman, David and Underwood, Ted and Smith, Noah A.},
  booktitle={Proceedings of the 52nd Annual Meeting of the Association for Computational Linguistics (ACL)},
  pages={370--379},
  year={2014}
}

@article{van2014extended,
  title={The Extended Transportation-Imagery Model: A Meta-Analysis of the Antecedents and Consequences of Consumers' Narrative Transportation},
  author={Van Laer, Tom and De Ruyter, Ko and Visconti, Luca M. and Wetzels, Martin},
  journal={Journal of Consumer Research},
  volume={40},
  number={5},
  pages={797--817},
  year={2014}
}

@inproceedings{maharjan2018letting,
  title={Letting Emotions Flow: Success Prediction by Modeling the Flow of Emotions in Books},
  author={Maharjan, Suraj and Kar, Sudipta and Montes, Manuel and Gonz{\'a}lez, Fabio A. and Solorio, Thamar},
  booktitle={Proceedings of the 2018 Conference of the North American Chapter of the Association for Computational Linguistics: Human Language Technologies (NAACL-HLT)},
  pages={259--265},
  year={2018}
}

@article{manski2004measuring,
  title={Measuring Expectations},
  author={Manski, Charles F.},
  journal={Econometrica},
  volume={72},
  number={5},
  pages={1329--1376},
  year={2004}
}

@article{nair2007intertemporal,
  title={Intertemporal Price Discrimination with Forward-Looking Consumers: Application to the {US} Market for Console Video-Games},
  author={Nair, Harikesh},
  journal={Quantitative Marketing and Economics},
  volume={5},
  number={3},
  pages={239--292},
  year={2007}
}

@article{caplin2001psychological,
  title={Psychological Expected Utility Theory and Anticipatory Feelings},
  author={Caplin, Andrew and Leahy, John},
  journal={Quarterly Journal of Economics},
  volume={116},
  number={1},
  pages={55--79},
  year={2001}
}

@article{simonov2023suspense,
  title={Suspense and Surprise in Media Product Design: Evidence from {Twitch.tv}},
  author={Simonov, Andrey and Ursu, Raluca M. and Zheng, Carolina},
  journal={Journal of Marketing Research},
  volume={60},
  number={1},
  pages={1--24},
  year={2023}
}

@techreport{radford2019language,
  title={Language Models are Unsupervised Multitask Learners},
  author={Radford, Alec and Wu, Jeffrey and Child, Rewon and Luan, David and Amodei, Dario and Sutskever, Ilya},
  institution={OpenAI},
  year={2019}
}

@article{goli2024capturing,
  title={Frontiers: Can Large Language Models Capture Human Preferences?},
  author={Goli, Ali and Singh, Amandeep},
  journal={Marketing Science},
  volume={43},
  number={4},
  pages={709--722},
  year={2024}
}

@article{brand2023using,
  title={Using {LLMs} for Market Research},
  author={Brand, James and Israeli, Ayelet and Ngwe, Donald},
  journal={Harvard Business School Marketing Unit Working Paper},
  number={23-062},
  year={2023}
}

@article{arora2025hybrid,
  title={{AI}--Human Hybrids for Marketing Research: Leveraging Large Language Models ({LLMs}) as Collaborators},
  author={Arora, Neeraj and Chakraborty, Ishita and Nishimura, Yohei},
  journal={Journal of Marketing},
  volume={89},
  number={2},
  pages={43--70},
  year={2025}
}

@inproceedings{filippas2024large,
  title={Large Language Models as Simulated Economic Agents: What Can We Learn from Homo Silicus?},
  author={Horton, John J. and Filippas, Apostolos and Manning, Benjamin S.},
  booktitle={Proceedings of the 25th ACM Conference on Economics and Computation (EC)},
  year={2024}
}

@article{bybee2023surveying,
  title={Surveying Generative AI's Economic Expectations},
  author={Bybee, Leland},
  journal={arXiv preprint arXiv:2305.02823},
  year={2023}
}

@article{mohammad2013crowdsourcing,
  title={Crowdsourcing a word--emotion association lexicon},
  author={Mohammad, Saif M and Turney, Peter D},
  journal={Computational intelligence},
  volume={29},
  number={3},
  pages={436--465},
  year={2013},
  publisher={Wiley Online Library}
}

@article{bisbee2024synthetic,
  title={Synthetic Replacements for Human Survey Data? The Perils of Large Language Models},
  author={Bisbee, James and Clinton, Joshua D. and Dorff, Cassy and Kenkel, Brenton and Larson, Jennifer M.},
  journal={Political Analysis},
  volume={32},
  number={4},
  pages={401--416},
  year={2024},
  publisher={Cambridge University Press}
}

@article{deng2026examining,
  title={Examining and Addressing Barriers to Diversity in LLM-Generated Ideas},
  author={Deng, Yuting and Brucks, Melanie and Toubia, Olivier},
  journal={arXiv preprint arXiv:2602.20408},
  year={2026},
  doi={10.48550/arXiv.2602.20408}
}

@article{egami2022make,
  title={How to make causal inferences using texts},
  author={Egami, Naoki and Fong, Christian J and Grimmer, Justin and Roberts, Margaret E and Stewart, Brandon M},
  journal={Science Advances},
  volume={8},
  number={42},
  pages={eabg2652},
  year={2022},
  publisher={American Association for the Advancement of Science}
}

@inproceedings{veitch2020adapting,
  title={Adapting text embeddings for causal inference},
  author={Veitch, Victor and Sridhar, Dhanya and Blei, David},
  booktitle={Conference on uncertainty in artificial intelligence},
  pages={919--928},
  year={2020},
  organization={PMLR}
}

@article{dube2010tipping,
  title={Tipping and concentration in markets with indirect network effects},
  author={Dub{\'e}, Jean-Pierre H and Hitsch, G{\"u}nter J and Chintagunta, Pradeep K},
  journal={Marketing Science},
  volume={29},
  number={2},
  pages={216--249},
  year={2010},
  publisher={INFORMS}
}

@article{ching2020identification,
  title={Identification and estimation of forward-looking behavior: The case of consumer stockpiling},
  author={Ching, Andrew T and Osborne, Matthew},
  journal={Marketing Science},
  volume={39},
  number={4},
  pages={707--726},
  year={2020},
  publisher={INFORMS}
}

@article{feder2022causal,
  title={Causal inference in natural language processing: Estimation, prediction, interpretation and beyond},
  author={Feder, Amir and Keith, Katherine A and Manzoor, Emaad and Pryzant, Reid and Sridhar, Dhanya and Wood-Doughty, Zach and Eisenstein, Jacob and Grimmer, Justin and Reichart, Roi and Roberts, Margaret E and others},
  journal={Transactions of the Association for Computational Linguistics},
  volume={10},
  pages={1138--1158},
  year={2022},
  publisher={MIT Press One Broadway, 12th Floor, Cambridge, Massachusetts 02142, USA~…}
}

@article{manzoor2023designing,
  title={Designing Effective Music Excerpts},
  author={Manzoor, Emaad and Malik, Nikhil},
  journal={arXiv preprint arXiv:2309.14475},
  year={2023}
}

@article{wei2025unstructured,
  title={Unstructured data, econometric models, and estimation bias},
  author={Wei, Yanhao'Max' and Malik, Nikhil},
  journal={USC Marshall School of Business Research Paper Sponsored by iORB},
  year={2025}
}

@inproceedings{tian2024narratives,
  title={Are Large Language Models Capable of Generating Human-Level Narratives?},
  author={Tian, Yufei and Huang, Tenghao and Liu, Miri and Jiang, Derek and Spangher, Alexander and Chen, Muhao and May, Jonathan and Peng, Nanyun},
  booktitle={Proceedings of the 2024 Conference on Empirical Methods in Natural Language Processing (EMNLP)},
  year={2024}
}
